\documentclass[letterpaper, 10 pt, conference]{ieeeconf}  % Comment this line out if you need a4paper

\IEEEoverridecommandlockouts                              % This command is only needed if 
\usepackage{graphicx} % for pdf, bitmapped graphics files
\usepackage{epsfig} % for postscript graphics files
\usepackage{mathptmx} % assumes new font selection scheme installed
\usepackage{times} % assumes new font selection scheme installed
\usepackage{amsmath} % assumes amsmath package installed
\usepackage{amssymb}  % assumes amsmath package installed
\usepackage{tikz}

\usepackage{mathtools}
\usepackage{epstopdf}
\usepackage{tikz}
\usepackage{algorithm}
\usepackage{algpseudocode}
\usepackage{amsfonts}
\usepackage{caption}
\usepackage{subcaption}
\usepackage{dblfloatfix}
\usepackage{fixltx2e}
\usepackage{calrsfs}
\usepackage{dsfont}

\DeclareMathAlphabet{\pazocal}{OMS}{zplm}{m}{n}

\usepackage[T1]{fontenc}

\newcommand{\coms}{\mathbf{r}}
\newcommand{\moms}{\mathbf{h}}
\newcommand{\lmoms}{\mathbf{l}}
\newcommand{\amoms}{\mathbf{k}}
\newcommand{\forcevars}{\mathbf{f}}
\newcommand{\copvars}{\mathbf{z}}

\newcommand{\norm}[1]{\left\lVert#1\right\rVert}

\renewcommand{\baselinestretch}{0.95}

\title{\LARGE \bf On Time Optimization of Centroidal Momentum Dynamics }

\author{Brahayam Ponton$^{1}$, Alexander Herzog$^{1}$, Andrea Del Prete$^{1}$, Stefan Schaal$^{1,2}$ and Ludovic Righetti$^{1,3}$% <-this % stops a space
	\thanks{This research was supported by New York University, the Max-Planck
		Society and the European Union's Horizon 2020 research and innovation
		programme (grant agreement No 780684 and European Research Council's
		grant No 637935). We would like to thank the reviewers for valuable comments on the first version of this article.}% <-this % stops a space
	\thanks{$^{1}$Max Planck Institute for Intelligent Systems, Tuebingen, Germany}
	\thanks{$^{2}$University of Southern California, Los Angeles, USA}
	\thanks{$^{3}$New York University, New York, USA}
}

\usetikzlibrary{calc}
\usetikzlibrary{mindmap,shapes,arrows}

\definecolor{lightblue}{rgb}{0.145,0.6666,1}

\newlength{\FSZ}
\newcommand{\drawvideo}[3]{% [0 0.25 0.5 0.75 1 1.25 1.5]
	\noindent\pgfmathsetlength{\FSZ}{\linewidth/#2}
	\begin{tikzpicture}[outer sep=0pt,inner sep=0pt,x=\FSZ,y=\FSZ]
	\draw[color=lightblue!50!black] (0,0) node[outer sep=0pt,inner sep=0pt,text width=\linewidth,minimum height=0] (video) {\noindent#3};
	\path [fill=lightblue!50!black,line width=0pt] 
	(video.north west) rectangle ([yshift=\FSZ] video.north east) 
	\foreach \x in {1,2,...,#2} {
		{[rounded corners=0.6] ($(video.north west)+(-0.7,0.8)+(\x,0)$) rectangle +(0.4,-0.6)}
	}
	;
	\path [fill=lightblue!50!black,line width=0pt] 
	([yshift=-1\FSZ] video.south west) rectangle (video.south east) 
	\foreach \x in {1,2,...,#2} {
		{[rounded corners=0.6] ($(video.south west)+(-0.7,-0.2)+(\x,0)$) rectangle +(0.4,-0.6)}
	}
	;
	\foreach \x in {1,...,#1} {
		\draw[color=lightblue!50!black] ([xshift=\x\linewidth/#1] video.north west) -- ([xshift=\x\linewidth/#1] video.south west);
	}
	\foreach \x in {0,#1} {
		\draw[color=lightblue!50!black] ([xshift=\x\linewidth/#1,yshift=1\FSZ] video.north west) -- ([xshift=\x\linewidth/#1,yshift=-1\FSZ] video.south west);
	}
	\end{tikzpicture}
}

\begin{document}

\maketitle
\thispagestyle{empty}
\pagestyle{empty}

%%%%%%%%%%%%%%%%%%%%%%%%%%%%%%%%%%%%%%%%%%%%%%%%%%%%%%%%%%%%%%%%%%%%
\renewcommand{\thefootnote}{\fnsymbol{footnote}}
\begin{abstract}
Recently, the centroidal momentum dynamics has received substantial attention to plan dynamically consistent motions for robots with arms and legs in multi-contact scenarios. However, it is also non convex which renders any optimization approach difficult and timing is usually kept fixed in most trajectory optimization techniques to not introduce additional non convexities to the problem. But this can limit the versatility of the algorithms. In our previous work, we proposed a convex relaxation of the problem that allowed to efficiently compute momentum trajectories and contact forces. However, our approach could not minimize a desired angular momentum objective which seriously limited its applicability. Noticing that the non-convexity introduced by the time variables is of similar nature as the centroidal dynamics one, we propose two convex relaxations to the problem based on trust regions and soft constraints. The resulting approaches can compute time-optimized dynamically consistent trajectories sufficiently fast to make the approach realtime capable. The performance of the algorithm is demonstrated in several multi-contact scenarios for a humanoid robot. In particular, we show that the proposed convex relaxation of the original problem finds solutions that are consistent with the original non-convex problem and illustrate how timing optimization allows to find motion plans that would be difficult to plan with fixed timing \footnote[2]{Implementation details and demos can be found in the source code available at https://git-amd.tuebingen.mpg.de/bponton/timeoptimization.}.
\end{abstract}
%\vspace{-0.1cm}
%
%%%%%%%%%%%%%%%%%%%%%%%%%%%%%%%%%%%%%%%%%%%%%%%%%%%%%%%%%%%%%%%%%%%%
\section{INTRODUCTION}
Motion optimization for robots with arms and legs such as humanoids is a challenging task for many reasons including very-high dimensionality, problem discontinuities due to intermittent contacts, non convex optimization landscapes prone to local minima, realtime constraints to find a solution, quality of the solution for execution on a real robot, etc. 
Yet, these challenges have inspired researchers for many years to develop optimization methods that show really impressive simulation results  \cite{DBLP:conf/humanoids/DaiVT14, Herzog-2016b, journals/tog/MordatchTP12, DBLP:conf/wafr/PosaT12, DBLP:conf/iros/TassaET12}. 

On the more practical side however, the most successful methods have not been the most complex and computationally expensive ones, but the ones that provide sufficient flexibility to perform a desired task while being well suited for model predictive control \cite{JustinMomentumOptimization, journals/trob/EnglsbergerOA15, DBLP:conf/humanoids/SherikovDW14, conf/humanoids/Wieber06, Caron:2016wt, Audren:2014gl}.
Indeed the ability to compute motions in a receding horizon fashion is very important to provide
the necessary reactivity to the robot behavior in uncertain environments.

In recent years, the centroidal momentum dynamics model \cite{Orin:2013ey,Kajita:2003gj} has become a popular model for multi-contact dynamic full-body motions \cite{DBLP:conf/humanoids/DaiVT14,Wensing:2013fm}. Indeed, this model, under the assumption of enough torque authority, provides sufficient conditions for planning dynamically feasible motions \cite{AlexHumanoidsPaper,TROCarpentier}, and is simple enough such that the problem could be solved with close to realtime rates \cite{JustinMomentumOptimization}.
Notably, it was shown in \cite{TROCarpentier} that such plans could be successfully used on a humanoid robot.

However, the centroidal momentum dynamics is not convex, which renders the optimization problem difficult.
Recent works have looked at the mathematical structure of the problem to find
more efficient optimization algorithms.
In \cite{TROCarpentier}, a multiple-shooting method is used to efficiently optimize the centroidal dynamics.
In \cite{Dai:2016hz} a convex bound on the rate of change in angular momentum is used to minimize a worst-case bound on the $l_1$ norm of the angular momentum. 
In \cite{Herzog-2016b}, it was shown that the non convex part could be decomposed as a difference of
quadratic functions. This allows an efficient convex approximation of the problem that can be used in
sequential quadratic programming approaches. 
The paper also proposed a method to efficiently compute dynamically consistent full-body motions by alternating centroidal dynamics optimization with full robot kinematics optimization. 

In our previous work \cite{ConvexModelMomentumDynamics}, we proposed a convex relaxation
of the problem that allowed to find solutions at realtime rates. However, our approach was using
a proxy function to minimize the angular momentum: it was minimizing the sum of the squares of the 
quadratic functions composing the non-convex part of the equations.
Thus, it was not possible to include an explicit target momentum in the cost function, which 
limits the space of desirable solutions where momentum is effectively minimized; which made this approach not well suited to be used directly in the alternating full-body optimization method proposed in \cite{Herzog-2016b}. The new approach presents a more general approximation of the nonconvex constraints, allowing to also investigate an under-studied aspect namely the importance of timing for centroidal momentum optimization.
Several works have realized the importance of including time as an optimization variable \cite{DBLP:conf/humanoids/DaiVT14, DBLP:conf/wafr/PosaT12, JustinMomentumOptimization, TOPP, TimeSwitchedJonas} and in doing so have shown great simulation and experimental results. However, including time optimization is usually computationally very costly due to the non convexity introduced in the discretized dynamics.

In this paper, we propose two methods for convex relaxation of the centroidal momentum dynamics optimization problem that allows to include an explicit angular momentum objective. 
Moreover, noticing that the non-convexity introduced by the time variable is of similar nature as the torque cross product allows us to use the same relaxation approaches for time optimization.
The resulting algorithm allows to compute dynamically consistent plans close to realtime.
Experiments demonstrate the computational efficiency of our approach in multi-contact scenarios.
In particular, we show that the numerical solutions found in our method are very close to the original dynamics (measured by the amount of constraint violation), suggesting that the convex relaxation is a good approximation of the original problem. Then we show that optimizing time allows to find solutions that could not be easily found with fixed time optimization.
We also combine our approach with the alternating approach proposed in \cite{Herzog-2016b}
to compute dynamically consistent full-body motions.

The remainder of this paper is structured as follows. In Sec. \ref{sec:problem_formulation}, we present the problem formulation. Then, in Sec. \ref{sec:technical_approach}, we show how to efficiently find a solution to the centroidal momentum dynamics problem. We show experimental results in Sec. \ref{sec:experiments} and conclude the paper in Sec. \ref{sec:conclusion}.
%
%%%%%%%%%%%%%%%%%%%%%%%%%%%%%%%%%%%%%%%%%%%%%%%%%%%%%%%%%%%%%%%%%%%%
\section{PROBLEM FORMULATION} \label{sec:problem_formulation}
The equations of motion that describe the dynamic evolution of a floating-base rigid body system are given by
\begin{equation*}
   \textbf{M}(\textbf{q})\ddot{\textbf{q}} + \textbf{N}(\textbf{q},\dot{\textbf{q}}) = \textbf{S}^{T} \tau_{j} + \textbf{J}_{\textrm{e}}^{T} \lambda \enspace ,
\end{equation*}
\noindent where the robot state is denoted by $\textbf{q}=\begin{bmatrix} x^{T} & q_{j}^{T} \end{bmatrix}^{T}$, and comprises the position and orientation of a floating base frame in the robot relative to an inertial frame $x \in SE(3)$, and the joint configuration $q_{j} \in \mathbb{R}^{n_{j}}$. The inertia matrix is denoted by $\textbf{M}(\textbf{q}) \in \mathbb{R}^{(n_{j}+6) \times (n_{j}+6)}$, the vector of nonlinear terms $\textbf{N}(\textbf{q},\dot{\textbf{q}}) \in \mathbb{R}^{n_{j}+6}$  includes Coriolis, centrifugal, gravity and friction forces, the selection matrix $\mathbf{S}=\begin{bmatrix} 0^{n_{j} \times 6} & I^{n_{j} \times n_{j}} \end{bmatrix}$ represents the system under-actuation, namely that $x \in SE(3)$ is not directly actuated by the vector of joint torques $\tau_{j} \in \mathbb{R}^{n_{j}}$, but indirectly through a vector of generalized forces $\lambda$ and the Jacobian of the contact constraints $\mathrm{J}_{\mathrm{e}}$.

The system under-actuation leads to a dynamics decomposition into an actuated (subscript ${a}$) and un-actuated parts (subscript $u$) as follows:
\begin{subequations}
\begin{align}
   \textbf{M}_{\mathrm{a}}(\textbf{q})\ddot{\textbf{q}} + \textbf{N}_{\mathrm{a}}(\textbf{q},\dot{\textbf{q}}) &= \tau_{j} + \textbf{J}_{\mathrm{e,a}}^{T} \lambda \label{eq_actuated_part}\\
   \textbf{M}_{\mathrm{u}}(\textbf{q})\ddot{\textbf{q}} + \textbf{N}_{\mathrm{u}}(\textbf{q},\dot{\textbf{q}}) &= \textbf{J}_{\mathrm{e,u}}^{T} \lambda \label{eq_newton_euler}
\end{align}
\end{subequations}
Equation \eqref{eq_newton_euler}, known as the Newton-Euler equations, tells us that the systems' change of momentum depends on external contact forces. Any combination of forces $\lambda$ and accelerations $\ddot{\textbf{q}}$ can be realized, if they are consistent with the underactuated dynamics \eqref{eq_newton_euler}, and there is enough torque authority \eqref{eq_actuated_part} \cite{AlexAuroPaper, WieberNonholonomy}. This natural decomposition suggests that satisfaction of the momentum equation \eqref{eq_newton_euler} is sufficient to guarantee dynamic feasibility, and equation \eqref{eq_actuated_part} ensures kinematic feasibility and torque limits.
\subsection{Dynamics Model}
As a consequence of the last observation, under the assumption of enough torque authority and kinematic reachability, a necessary condition for planning physically consistent motions is that the total wrench generated by external and gravitational forces \eqref{eqns_centroidal_momentum_dynamics} equals the rate of momentum computed from the robot joint angles and velocities \eqref{kin_centroidal_momentum}  \cite{DBLP:conf/humanoids/DaiVT14}. On the one hand, the centroidal momentum, computed from external forces, expressed at the robot center of mass is
\begin{equation}
{\dot{\moms}} =
\begin{bmatrix}
{\dot{\coms}}  \\[0.2em]
{\dot{\lmoms}}  \\[0.2em]
{\dot{\amoms}} \\[0.2em]
\end{bmatrix} = 
\begin{bmatrix}
\frac{1}{m} {\lmoms}            \\[0.2em]
m \mathbf{g} + \sum_{\text{e}} \forcevars_{\text{e}}  \\[0.2em]
\sum_{\text{e}} (\mathbf{p}_{\text{e}}+ \mathbf{R}^{\mathrm{x,y}}_{\mathrm{e}} \copvars_{\text{e}}-{\coms}) \times \forcevars_{\text{e}} + \mathbf{R}^{\mathrm{z}}_{\mathrm{e}} \mathbf{\tau}_{\text{e}} \\[0.2em]
\end{bmatrix}  \label{eqns_centroidal_momentum_dynamics}
\end{equation}
\noindent where $\coms$ denotes the center of mass position, $\lmoms$ the linear and $\amoms$ the angular momentum. $\moms$ is a shortcut vector comprising $\coms$, $\lmoms$ and $\amoms$. The total robot mass is $m$ and $\mathbf{g}$ the gravity vector. The position of the $e$ end-effector is denoted $\mathbf{p}_{\text{e}}$, $\copvars_{\text{e}} \in \mathbb{R}^{2}$ is the center of pressure (CoP) expressed in local end-effector coordinates. $\mathbf{R}^{\mathrm{x,y}}_{\mathrm{e}} \in \mathbb{R}^{3 \times 2}$ represents the first two columns of the rotation matrix $\mathbf{R}_{\mathrm{e}} \in \mathbb{R}^{3 \times 3}$ that maps quantities from end-effector frame to the inertial coordinate frame. $\forcevars_{\text{e}} \in \mathbb{R}^{3}$ and  $\mathbf{R}^{\mathrm{z}}_{\mathrm{e}} \tau_{\mathrm{e}} \in \mathbb{R}^{3}$ are forces and torques acting at contact point $\mathbf{p}_{\text{e}}+ \mathbf{R}^{\mathrm{x,y}}_{\mathrm{e}} \copvars_{\text{e}}$, represented in inertial frame. $\tau_{\mathrm{e}} \in \mathbb{R}$ is the torque around the $z$, upward pointing, axis expressed in end-effector frame. $\mathbf{R}^{\mathrm{z}}_{\mathrm{e}}$ maps $\tau_{\mathrm{e}}$ to the inertial coordinate frame.

On the other hand, the centroidal momentum, computed from the robot joint angles and velocities, is given by \cite{OrinCentroidalMomentum}
\begin{equation}
\begin{bmatrix}
\lmoms  \\[0.2em]
\amoms
\end{bmatrix} = 
\mathbf{A}(\mathbf{q}) \dot{\mathbf{q}}
\label{kin_centroidal_momentum}
\end{equation}
\noindent where $\mathbf{A}(\mathbf{q}) \in \mathbb{R}^{6 \times n_{j}+6}$ is the centroidal momentum matrix. we highlight that the rate of change of momentum, as given by \eqref{eqns_centroidal_momentum_dynamics}, only depends on dynamic quantities, while the momentum, as given by \eqref{kin_centroidal_momentum}, only depends on kinematic quantities. This separation, suggested in \cite{Herzog-2016b, DBLP:conf/humanoids/DaiVT14}, allows the use of an iterative procedure, where one can alternate between a kinematic and a dynamic optimization to solve the joint problem. Both optimization procedures need only agree on the center of mass trajectory, momentum and contact locations and such agreement is enforced by a cost function in each optimization algorithm.
%For instance, the dynamics optimization will have as objective tracking a momentum trajectory that was optimized by the kinematic optimization, while satisfying dynamic constraints; and the kinematics optimization will have as objective tracking a momentum trajectory, result of a dynamics optimization, while satisfying kinematic constraints. 
The benefit of this separation is that the optimization problems can be solved more easily separately than as a joint problem and the inherent structure of each problem can be exploited in dedicated solvers.
\subsection{Trajectory Optimization}
In this paper we focus on the centroidal dynamics optimization problem and only briefly comment on
the alternating full-body optimization procedure as we will use it in the experiments section.
First, we discretize the differential equations \eqref{eqns_centroidal_momentum_dynamics}-\eqref{kin_centroidal_momentum} into algebraic equations and then use the iterative procedure 
described in \cite{Herzog-2016b} to alternate between a dynamics and a kinematics optimization.
\subsubsection{Kinematics Optimization Problem}
We will not focus on the kinematic optimization. However, we would like to mention that we use at each time step an inverse kinematics procedure, whose objective is to track a desired center of mass position, linear and angular momenta, regularize rates of momentum, track desired motions for unconstrained end-effectors, regularize joint posture towards a default posture and regularize joint velocities and accelerations. The constraints include the evolution of linear and angular momenta using the centroidal momentum dynamics \eqref{kin_centroidal_momentum}, evolution of center of mass according to linear momentum, evolution of endeffectors based on the end-effector jacobians, joint limits and constraints for active end-effectors.
\subsubsection{Dynamics Optimization Problem}
We are interested in the optimization of dynamic motions including momentum trajectories, contact forces and timings, under the non-convex and non-linear centroidal momentum dynamics, which could later be realized by a low-level controller such as an inverse dynamics one \cite{AlexHumanoidsPaper}. We assume a fixed set of contacts, however they can be easily included as optimization variables constrained to lie within the stepping stones \cite{DBLP:conf/humanoids/DeitsT14, HyQWalking}. Formally, the objective to be minimized in our optimization problem is:
\begin{equation}
\min_{ \forcevars_{\text{e}}, \tau_{\mathrm{e}}, \copvars_{\text{e}}, \Delta_{t} } \quad \phi_{N}(\moms_{n}-\moms_{n_{\mathrm{des}}}) + \sum\limits_{t=1}^{n-1} \ell_{t}(\moms - \moms_{\mathrm{des}}, \copvars_{\text{e}}, \forcevars_{\text{e}}, \tau_{\mathrm{e}}, \Delta_{t}) 
\label{eq_generic_cost}
\end{equation}
More specifically, we would like to minimize a terminal cost $\phi_{N}(\moms_{n}-\moms_{n_{\mathrm{des}}})$ that penalizes the difference between the final state $\moms_{n}$ and the desired final state $\moms_{n_{\mathrm{des}}}$, and a running cost $\ell_{t}(\moms - \moms_{\mathrm{des}}, \copvars_{\text{e}}, \forcevars_{\text{e}}, \tau_{\mathrm{e}}, \Delta_{t})$, that penalizes the tracking performance of a desired linear and angular momentum trajectories $\moms - \moms_{\mathrm{des}}$, and regularizes the available controls, namely, forces $\forcevars_{\text{e}}$, torques $\tau_{\mathrm{e}}$ and time discretizations $\Delta_{t}$.

Desired momentum trajectories could be as trivial as zeros or could for instance come from the kinematic optimization.
%
%everywhere, or could for instance come  from a kinematic optimization, also capable of tracking momentum, but that can also more easily handle other type of constraints, such as obstacle avoidance, joint limits, among others, that if used in the dynamic optimization problem would render it more difficult to solve \cite{Herzog-2016b}.
%
The constraints of the optimization problem include a discrete form of the centroidal momentum dynamics \eqref{eqns_centroidal_momentum_dynamics}:
\begin{equation}
\moms =
\begin{bmatrix}
    \coms_{\text{t}}        \\[0.2em]
    \lmoms_{\text{t}}        \\[0.2em]
    \amoms_{\text{t}}        \\[0.2em]
    \dot{\lmoms}_{\text{t}}  \\[0.2em]
    \dot{\amoms}_{\text{t}}  \\[0.2em]
 \end{bmatrix} = 
 \begin{bmatrix}
     \coms_{\text{t}-1} + \frac{1}{m} \lmoms_{\text{t}} \Delta_{\text{t}}  \\[0.2em]
     \lmoms_{\text{t}-1} + \dot{\lmoms}_{\text{t}} \Delta_{\text{t}}         \\[0.2em]
     \amoms_{\text{t}-1} + \dot{\amoms}_{\text{t}} \Delta_{\text{t}}       \\[0.2em]
     m \mathbf{g} + \sum_{\text{e}} \forcevars_{\text{e,t}}                    \\[0.2em]
     \sum_{\text{e}} \mathbf{\kappa_{\text{e,t}}}                                   \\[0.2em]
  \end{bmatrix} \label{eq_linear_momentum}
\end{equation}
\noindent The variable $\mathbf{\kappa_{\text{e,t}}}$ (end-effector contribution to angular momentum rate $\dot{\amoms}_{\text{t}}$) has been defined as
\begin{align}
\mathbf{\kappa_{\text{e,t}}} &= (\mathbf{p}_{\text{e,t}} + \mathbf{R}^{\mathrm{x,y}}_{\mathrm{e,t}} \copvars_{\text{e,t}} - \coms_{\text{t}}) \times  \forcevars_{\text{e,t}} + \mathbf{R}^{\mathrm{z}}_{\mathrm{e,t}} \mathbf{\tau}_{\text{e,t}}  \nonumber \\[0.1em]
&= \boldsymbol{\ell}_{\text{e,t}} \times  \forcevars_{\text{e,t}} + \mathbf{R}^{\mathrm{z}}_{\mathrm{e,t}} \mathbf{\tau}_{\text{e,t}}  \nonumber \\[0.1em]
&=   \begin{bmatrix*}[r]
    0                               & -{\ell}^{\text{ z}}_{\text{e,t}} &  {\ell}^{\text{ y}}_{\text{e,t}}  \\[0.2em]
    {\ell}^{\text{ z}}_{\text{e,t}} &  0                               & -{\ell}^{\text{ x}}_{\text{e,t}}  \\[0.2em]
   -{\ell}^{\text{ y}}_{\text{e,t}} &  {\ell}^{\text{ x}}_{\text{e,t}} &  0                                \\[0.2em]
  \end{bmatrix*} 
  \begin{bmatrix}
    \forcevars^{\text{ x}}_{\text{e,t}} \\[0.2em]
    \forcevars^{\text{ y}}_{\text{e,t}} \\[0.2em]
    \forcevars^{\text{ z}}_{\text{e,t}} \\[0.2em]
  \end{bmatrix} + \mathbf{R}^{\mathrm{z}}_{\mathrm{e,t}} \mathbf{\tau}_{\text{e,t}} \label{eq_torque_amomrate}
\end{align}
\noindent where for simplicity of notation, we have introduced the change of variable ${\ell}_{\text{e,t}} = (\mathbf{p}_{\text{e,t}} + \mathbf{R}^{\mathrm{x,y}}_{\mathrm{e,t}} \copvars_{\text{e,t}} - \coms_{\text{t}})$. Physical constraints such as friction cone, CoP within region of support and torque limits are given by 
\begin{subequations}
	\begin{align}
	&\norm{\prescript{_\mathrm{L}}{}{\forcevars^{\text{ x}}_{\text{e,t}}}+\prescript{_\mathrm{L}}{}{\forcevars^{\text{ y}}_{\text{e,t}}}}_{2} \le \mu \prescript{_\mathrm{L}}{}{\forcevars^{\text{ z}}_{\text{e,t}}}, \hspace{0.4cm} \prescript{_\mathrm{L}}{}{\forcevars^{\text{ z}}_{\text{e,t}}} \ge 0, \label{eq_friction_cone} \\[0.3em]
	&\copvars^{\text{x,y}}_{\text{e,t}} \in \left[ \copvars^{\text{x,y}}_{\mathrm{min}}, \copvars^{\text{x,y}}_{\mathrm{max}} \right], \label{eq_conservative_cop}\\[0.3em]
	&\tau_{\text{e,t}} \in \left[ \tau_{\mathrm{min}}, \tau_{\mathrm{max}} \right] \label{eq_torque_cone} \\[0.3em]
	&\Delta_{t} \in \left[ \Delta_{\mathrm{min}}, \Delta_{\mathrm{max}} \right] \label{eq_time_bounds}  \\[0.3em]
	&\norm{\mathbf{p}_{\mathrm{e},t} - \coms_{\mathrm{t}}} \le \ell_{\mathrm{e}}^{\mathrm{max}} \label{eq_eef_len}
	\end{align}
	\label{eq_friction_cop_torque}
\end{subequations}
where $\prescript{_\mathrm{L}}{}{\forcevars_{\text{e,t}}} =  \mathbf{R}^{T}_{\mathrm{e,t}} \forcevars_{\text{e,t}}$ is the end-effector force in local frame. \eqref{eq_friction_cone} states that forces belong to a friction cone with friction coefficient $\mu$. Friction cones could alternatively be approximated by the usual pyramids as a polyhedral approximation. \eqref{eq_conservative_cop} expresses that the CoP should be within a conservative region with respect to the real physical available region. \eqref{eq_torque_cone} constrains the torque to a bounded region, \cite{DBLP:conf/icra/CaronPN15} also provides precise closed-form formulas for it under polyhedral approximation of the friction cone. Equation \eqref{eq_time_bounds} constrains the time discretization variable to a bounded region. Finally, equation \eqref{eq_eef_len} constrains the distance between the current position of the center of mass $\coms_{\mathrm{t}}$ and the end-effector contact point $\mathbf{p}_{\mathrm{e,t}}$ to be less than the maximum length of the end-effector $\ell_{\mathrm{e}}^{\mathrm{max}}$. It includes a constant offset, when the end-effectors of interest are the arms of the robot.
%
%\begin{subequations}
%	\begin{align}
%	&\norm{\prescript{_\mathrm{L}}{}{\forcevars^{\text{ x}}_{\text{e,t}}}+\prescript{_\mathrm{L}}{}{\forcevars^{\text{ y}}_{\text{e,t}}}}_{2} \le \mu \prescript{_\mathrm{L}}{}{\forcevars^{\text{ z}}_{\text{e,t}}}, \hspace{0.4cm} \prescript{_\mathrm{L}}{}{\forcevars^{\text{ z}}_{\text{e,t}}} \ge 0, \label{eq_friction_cone} \\[0.3em]
%	&\copvars^{\text{x,y}}_{\text{e,t}} \in \left[ \copvars^{\text{x,y}}_{\mathrm{min}}, \copvars^{\text{x,y}}_{\mathrm{max}} \right], \label{eq_conservative_cop}\\[0.3em]
%	&\tau_{\text{e,t}} \in \left[ \tau_{\mathrm{min}}, \tau_{\mathrm{max}} \right] \label{eq_torque_cone} \\[0.3em]
%	&\Delta_{t} \in \left[ \Delta_{\mathrm{min}}, \Delta_{\mathrm{max}} \right] \label{eq_time_bounds}  \\[0.3em]
%	&\norm{\mathbf{p}_{\mathrm{e},t} - \coms_{\mathrm{t}}} \le \ell_{\mathrm{e}}^{\mathrm{max}} \label{eq_eef_len}
%	\end{align}
%	\label{eq_friction_cop_torque}
%\end{subequations}

We would like to conclude this section by highlighting that the non-convexities of this problem are the bilinear terms of eq. \eqref{eq_linear_momentum}. For instance, terms where the time discretization variable $\Delta_{\text{t}}$ appears or the cross products of the angular momentum rate terms $\mathbf{\kappa_{\text{e,t}}}$. In traditional momentum optimization \cite{Herzog-2016b,TROCarpentier, caron2016humanoids}, the focus is on the cross products of the angular momentum, that introduce bilinear constraints to the optimization problem. It is important to note that including time discretizations as optimization variables introduces nonconvex constraints of the same bilinear nature. Therefore, the goal of the next section will be to devise methods that can approximate bilinear constraints, while still allowing us to efficiently find a solution to the optimization problem.
%
%%%%%%%%%%%%%%%%%%%%%%%%%%%%%%%%%%%%%%%%%%%%%%%%%%%%
%
\section{APPROACH} \label{sec:technical_approach}
In this section, we describe a tool to express bilinear constraints \eqref{eq_linear_momentum}-\eqref{eq_torque_amomrate} as a difference of convex functions, and then we show how to approximate these still nonconvex constraints, using the knowledge about their positive curvature.
\subsection{Disciplined Convex-Concave Programming}
In this subsection, we will describe a tool \cite{DBLP:conf/cdc/ShenDGB16} for handling bilinear constraints in optimization problems. The method decomposes a bilinear or nonconvex quadratic expression into a difference of convex functions; in other words, it decomposes the bilinear expression into a difference of two terms, each of which is a convex function. After the decomposition, the constraint will continue to be nonconvex; however, the terms composing it will have known curvature, which will let us perform an efficient approximation.%, where we exploit fast convergence properties of convex functions.

This approach, previously used and detailed in \cite{Herzog-2016b, ConvexModelMomentumDynamics}, analytically decomposes the nonconvex quadratic expressions of the angular momentum \eqref{eq_torque_amomrate} into a difference of convex (quadratic) functions. 
The set of difference of convex functions $\pazocal{C}^{\pm}$ is defined as:
\begin{align*}
\pazocal{C}^{\pm} = \bigg\{& \pazocal{C}^{+}(\mathbf{x}) - \pazocal{C}^{-}(\mathbf{x}) \;|\; \mathbf{x}\in\mathbb{R}^{n}, \big. \\
\big. &\pazocal{C}^{+}, \pazocal{C}^{-}: \mathbb{R}^{n} \rightarrow \mathbb{R}, \text{ are convex functions} \bigg\}
\end{align*}
Expressions such as scalar products $x^{T}y$ can be decomposed into a difference of convex functions, such as $Q^{+}-Q^{-}$
\begin{align*}
Q^{+} = \frac{1}{4} \norm{x+y}^{2} \text{, and } Q^{-} = \frac{1}{4} \norm{x-y}^{2}
\end{align*}
\noindent where $Q^{+} \in \pazocal{C}^{+}$ and $Q^{-} \in \pazocal{C}^{-}$ are convex functions. Using this idea, we can easily identify scalar products and transform quadratic expressions in cross products into scalar products, which can then be defined as elements of $\pazocal{C}^{\pm}$. As an example, we present the decomposition of a cross product:
\begin{align}
\ell \hspace{-0.05cm} \times \hspace{-0.05cm} \forcevars &=
\begin{bmatrix}
\underbrace{\begin{bmatrix} -{\ell}_{\text{z}} & \hspace{-0.2cm} {\ell}_{\text{y}} \end{bmatrix}}_{{\mathbf{a}_{\text{cvx}}^{T}}}
\overbrace{\begin{bmatrix} \forcevars_{\text{y}} \\  \forcevars_{\text{z}} \end{bmatrix}}^{{\mathbf{d}_{\text{cvx}}}}, \hspace{-0.3cm}&
\underbrace{\begin{bmatrix} {\ell}_{\text{z}} &  \hspace{-0.2cm} -{\ell}_{\text{x}} \end{bmatrix}}_{{\mathbf{b}_{\text{cvx}}^{T}}}
\overbrace{\begin{bmatrix} \forcevars_{\text{x}} \\  \forcevars_{\text{z}} \end{bmatrix}}^{{\mathbf{e}_{\text{cvx}}}}, \hspace{-0.3cm}&
\underbrace{\begin{bmatrix} -{\ell}_{\text{y}} & \hspace{-0.2cm} {\ell}_{\text{x}} \end{bmatrix}}_{{\mathbf{c}_{\text{cvx}}^{T}}}
\overbrace{\begin{bmatrix} \forcevars_{\text{x}} \\  \forcevars_{\text{y}} \end{bmatrix}}^{{\mathbf{f}_{\text{cvx}}}} 
\end{bmatrix}^{T}
\nonumber
\end{align}
\noindent and then each scalar product is defined as an element of $\pazocal{C}^{\pm}$.
\begin{align}
\ell \hspace{-0.05cm} \times \hspace{-0.05cm} \forcevars &= 
\frac{1}{4} 
\begin{bmatrix}
{\begin{Vmatrix} \mathbf{a}_{\text{cvx}} +  \mathbf{d}_{\text{cvx}} \end{Vmatrix}^{2}_{2} -
	\begin{Vmatrix} \mathbf{a}_{\text{cvx}} - \mathbf{d}_{\text{cvx}} \end{Vmatrix}^{2}_{2}} \\[0.15em]
{\begin{Vmatrix} \mathbf{b}_{\text{cvx}} +  \mathbf{e}_{\text{cvx}} \end{Vmatrix}^{2}_{2} -
	\begin{Vmatrix} \mathbf{b}_{\text{cvx}} - \mathbf{e}_{\text{cvx}} \end{Vmatrix}^{2}_{2}} \\[0.15em]
{\begin{Vmatrix} \mathbf{c}_{\text{cvx}} +  \mathbf{f}_{\text{cvx}} \end{Vmatrix}^{2}_{2} -
	\begin{Vmatrix} \mathbf{c}_{\text{cvx}} - \mathbf{f}_{\text{cvx}} \end{Vmatrix}^{2}_{2}} \\[0.15em]
\end{bmatrix}
\nonumber
\end{align}
For simplicity of presentation, we denote $\mathbf{p}=\mathbf{a}_{\text{cvx}}+\mathbf{d}_{\text{cvx}}$ and $\mathbf{q}=\mathbf{a}_{\text{cvx}}-\mathbf{d}_{\text{cvx}}$ ($\mathbf{p}, \mathbf{q} \in \mathbb{R}^{2}$). With this notation the first component of the torque contribution of an end-effector to the angular momentum rate dynamics \eqref{eq_torque_amomrate} becomes
\begin{align*}
\kappa^{x} &= \frac{1}{4} 
\begin{bmatrix}
\mathbf{p}^{T} \mathbf{p} - \mathbf{q}^{T} \mathbf{q}
\end{bmatrix} + \tau^{x} \enspace,
\end{align*}
which is equivalent to the following formulation
\begin{subequations}
	\begin{align}
	&\kappa^{x} = \frac{1}{4} 
	\begin{bmatrix}
	\bar{p} - \bar{q}
	\end{bmatrix} + \tau^{x} \nonumber \\
	&\bar{p} = \mathbf{p}^{T} \mathbf{p} \quad \bar{q} = \mathbf{q}^{T} \mathbf{q} \label{eq_known_curvature}
	\end{align}
\end{subequations}
\noindent where we have introduced the scalar variables $\bar{p}$, $\bar{q} \in \mathbb{R}_{+}$. Under this formulation the original nonconvex quadratic constraint has been now separated into a linear constraint, and two non-convex constraints \eqref{eq_known_curvature}. The difference is that, while in the original constraint, the hessian is an indefinite matrix, the hessian of the new constraints is positive semi-definite and therefore each term has known curvature.
\subsection{Approximations of Quadratic Equality Constraints}
In the last section, we have presented a method to analytically decompose a bilinear constraint into a linear constraint and two nonconvex quadratic equality constraints with known curvature, see \eqref{eq_known_curvature}. In this section, we present alternatives for dealing with each of the two nonconvex quadratic equality constraints with known curvature, namely, approximation using a trust region and use of soft constraints.
\begin{figure}[h]
	\centering
	\includegraphics[width=0.45\textwidth]{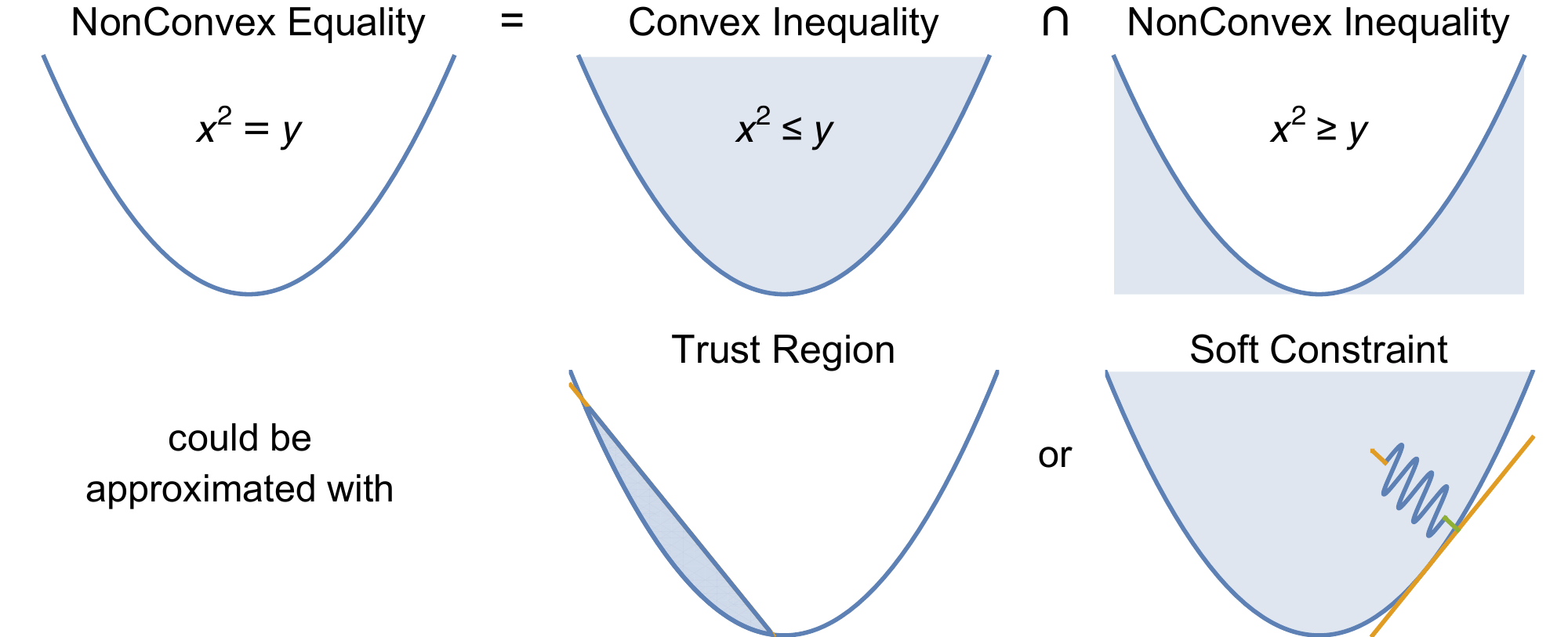}
	\caption{\small Alternative approximations of nonconvex quadratic equality constraint with known curvature: trust region and soft constraint.}
	\label{fig1:quadractic_equality_constraint}
	%\vspace{-0.4cm}
\end{figure}
\subsubsection{Approximation using a trust region}
A nonconvex quadratic equality constraint with known curvature can be thought of as the intersection of a convex and a nonconvex inequality constraints (Fig. \ref{fig1:quadractic_equality_constraint}). The approximation of the equality constraint using a trust region consists in keeping only the convex inequality constraint, but restricting its interior with a trust region. The effect of this, as can be seen in Fig. \ref{fig1:quadractic_equality_constraint} is that the search space is close to the boundary of the curve, and therefore the values of $y$ are close to $x^2$, as desired to approximate the original equality constraint.

Generalizing this example to the context of our problem, the nonconvex constraint \eqref{eq_known_curvature} can be equivalently represented as the intersection of a convex constraint $\bar{p} \succeq \mathbf{p}^{T}\mathbf{p}$ and a nonconvex constraint $\mathbf{p}^{T}\mathbf{p} \succeq \bar{p}$. The main idea of the method is to first obtain an initial guess of the optimal vector by solving the optimization problem using only the convex part of the search space, namely $\bar{p} \succeq \mathbf{p}^{T}\mathbf{p}$ (as in \cite{ConvexModelMomentumDynamics}), and then refine the solution by introducing the trust region. 

While the trust region could be introduced as a trivial box constraint with a threshold, that would constrain the value of ${\bar{p}}$ to values near $\mathbf{p}^{T}\mathbf{p}$, the best affine trust region that exploits the knowledge about the curvature of the function and the information contained in the current value of the optimal vector is a linear approximation such as $\mathbf{p}_{\textrm{val}}^{T}\mathbf{p}_{\textrm{val}} + 2 \mathbf{p}_{\textrm{val}} (\mathbf{p}-\mathbf{p}_{\textrm{val}})  \succeq {\bar{p}} - \sigma$, where $\sigma$ is a positive value representing a threshold, enough to provide a feasible interior to the intersection of the constraints, and $\mathbf{p}_{\textrm{val}}$ is any value taken by the variables $\mathbf{p}$ coming from the solution of the relaxed problem. Notice that if the hessian of the quadratic equality constraint where an indefinite matrix, this trust region would not constrain the problem as desired and instead lead to unbounded regions. The advantage of building the trust region this way is that $\sigma$ represents the maximum amount of desired constraint violation, which brings first  an automatic way to constrain the values of $\mathbf{p}$ around $\mathbf{p}_{\textrm{val}}$ that satisfy this amount, and second a method to further refine the solution by reducing the value of $\sigma$ (that increase approximation accuracy), as required by convergence tolerances.
 
While this approximation is still a relaxation of the constraint, it allows us to approximate the terms ${\bar{p}}$, ${\bar{q}}$, and consequently $\kappa^{x}$. This formulation has the advantage that we do not trade-off cost expressiveness, allowing e.g. quadratic terms over $\kappa^{x}$ without losing the convexity properties.
\subsubsection{Approximation using soft constraints}
This method is similar in spirit to the last one. It will first drop the nonconvex terms to find an initial guess for the optimal vector and then introduce cost heuristics, whose goal is to bias the solutions towards the boundaries. Unlike the previous method, this method does not restrict strictly the search space, but instead biases the solutions towards the boundaries by pulling the variables towards an underestimator of the function.

As in the previous case, the best affine underestimator, that exploits the knowledge about the curvature of the function and the current value of the optimal vector is a linearization $\mathbf{p}_{\textrm{val}}^{T}\mathbf{p}_{\textrm{val}} + 2 \mathbf{p}_{\textrm{val}} (\mathbf{p}-\mathbf{p}_{\textrm{val}})$. The heuristic is a quadratic term in the cost that penalizes the difference between the variable ${\bar{p}}$ and the linearization. This rewards the optimization for selecting values of ${\bar{p}}$ that are close to the boundaries of the constraint and are therefore feasible for the nonconvex constraint. As in the previous case, we do not trade-off expressiveness of the cost, allowing for example quadratic terms over $\kappa^{x}$ without losing the convexity properties.

\subsubsection{Convergence criteria}
The amount of constraint violation can be determined by how close the approximation is to the nonconvex constraint. In our setting, the convergence criteria is that the average error on center of mass and momentum trajectories, computed comparing the values of the corresponding variables coming from the relaxed optimization problem and the values computed integrating optimal torques and forces are below a certain residual error.

For problems involving time optimization, we have noticed that the speed of convergence of angular momentum (being the last to converge) to optimal values strongly depends on the convergence of first linear momentum and then center of mass. Based on this observation, we use in time optimization problems a first phase where the problem bilinear constraints are approximated as described in this section and let the algorithm discover a motion plan and its timing. In a second phase, we focus on the convergence of the approximated angular momentum, and do so by keeping the approximation of its cross products as described in this section and replacing the second order approximation of time related constraints with first order ones. This allows the optimizer to still make adjustments on the center of mass, linear momentum, and timings, while allowing us to speed up and increase accuracy of the convergence of angular momentum.
%
%%%%%%%%%%%%%%%%%%%%%%%%%%%%%%%%%%%%%%%%%%%%%%%%%%%%
%
\section{Experiments} \label{sec:experiments}
\begin{figure*}
	\centering
	\begin{subfigure}[b]{0.48\textwidth}   
		\centering
		\drawvideo{5}{80}{%
			\includegraphics[width=0.20\linewidth, trim={25cm 0cm 20cm 4cm}, clip]{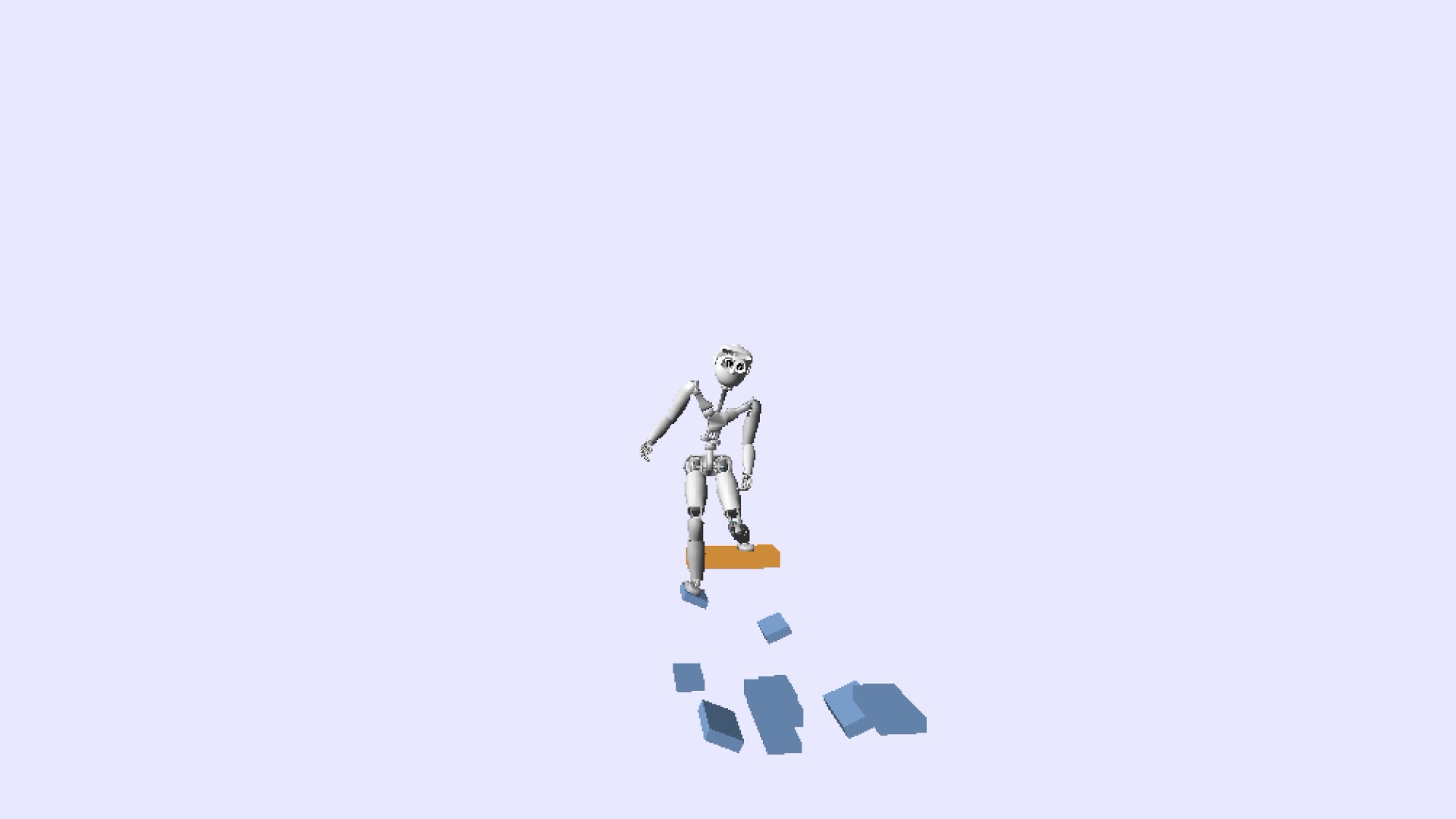}%
			\includegraphics[width=0.20\linewidth, trim={25cm 0cm 20cm 4cm}, clip]{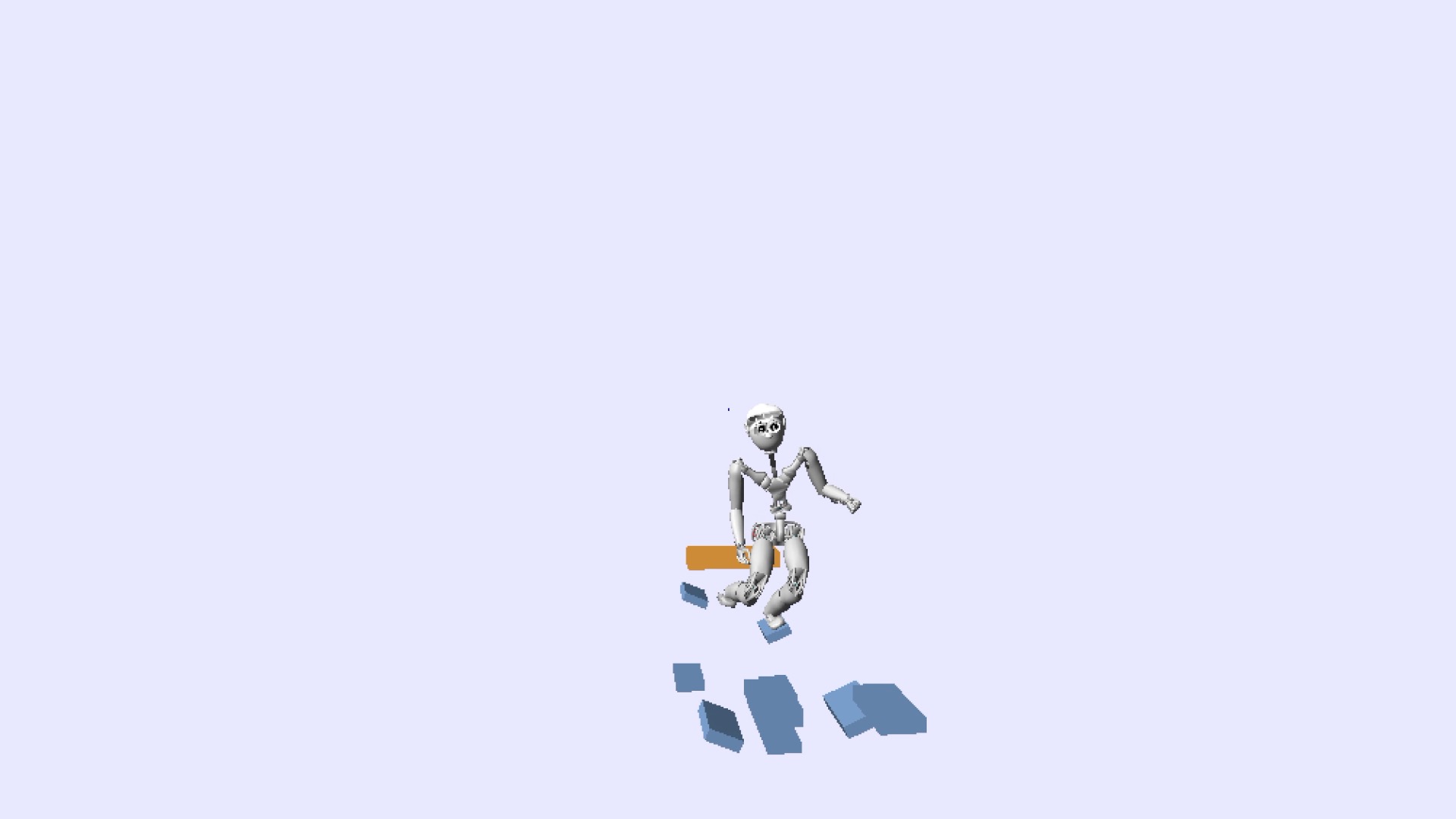}%
			\includegraphics[width=0.20\linewidth, trim={25cm 0cm 20cm 4cm}, clip]{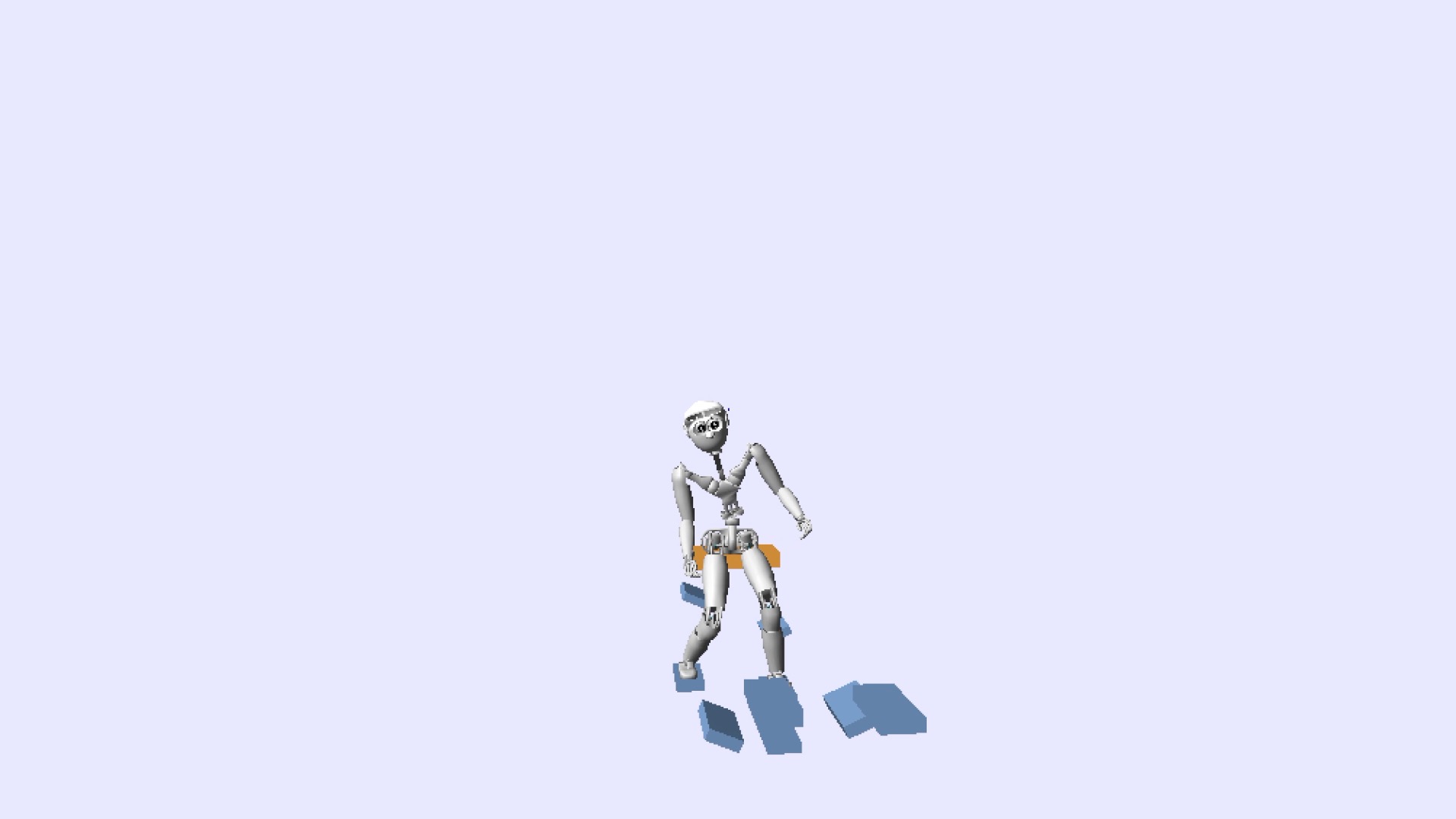}%
			\includegraphics[width=0.20\linewidth, trim={25cm 0cm 20cm 4cm}, clip]{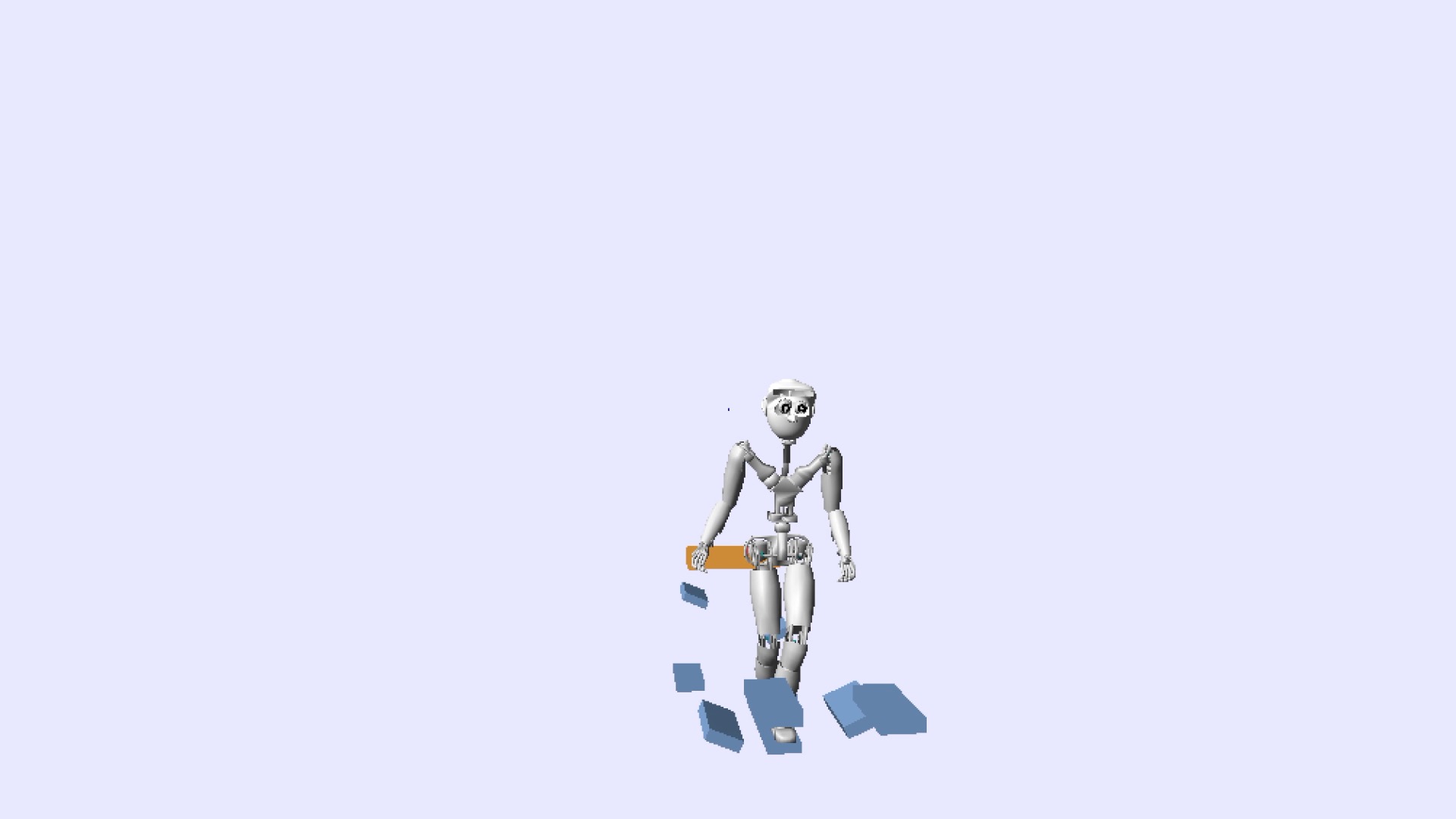}%
			\includegraphics[width=0.20\linewidth, trim={25cm 0cm 20cm 4cm}, clip]{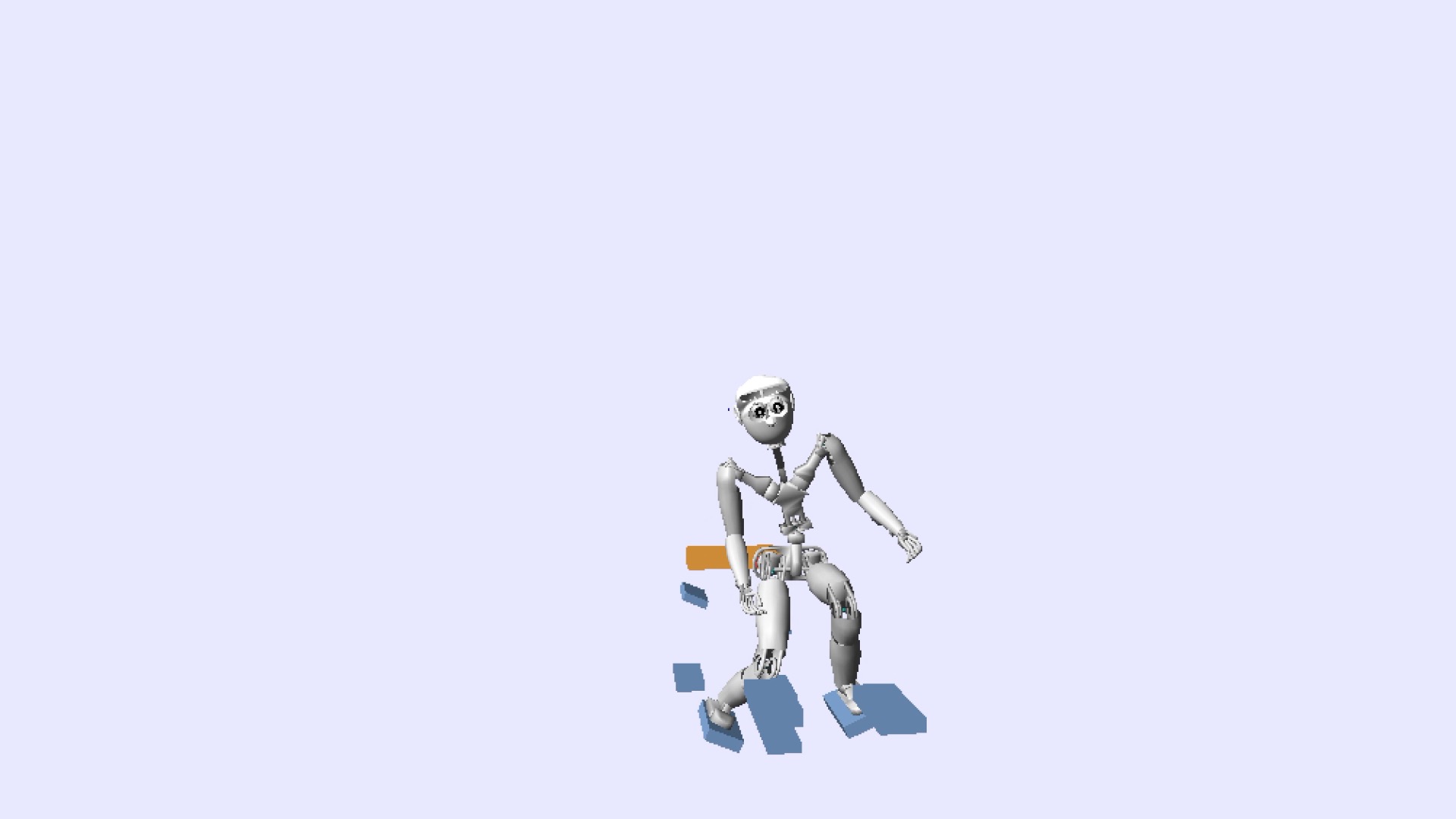}%
		}\\
		\vspace{0.4cm}
		\includegraphics[width=\textwidth]{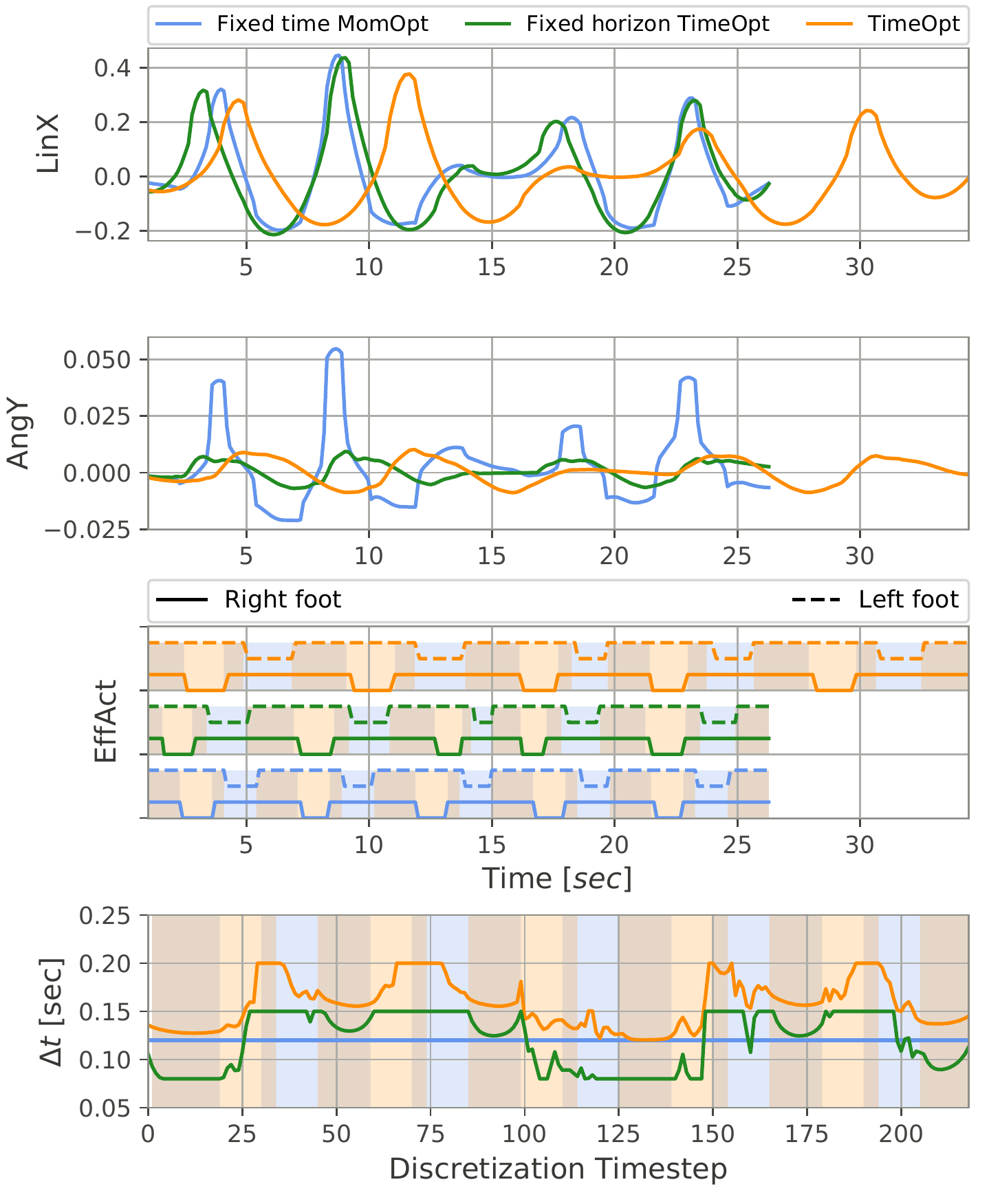} 
		\caption[]%
		{{\small This figure shows optimization results for the walking on uneven terrain motion optimizing only momentum (blue), including time in the optimization (orange) and time optimization with fixed time horizon (green). The number of kino-dynamic iterations is relative to the motion complexity and tuning effort, usually requiring 1 to 3 iterations.}}    
		\label{fig2:walking_motion}
	\end{subfigure}
	\quad
	\begin{subfigure}[b]{0.48\textwidth}   
		\centering
		\drawvideo{5}{80}{%
			\includegraphics[width=0.20\linewidth, trim={20cm 0cm 25cm 4cm}, clip]{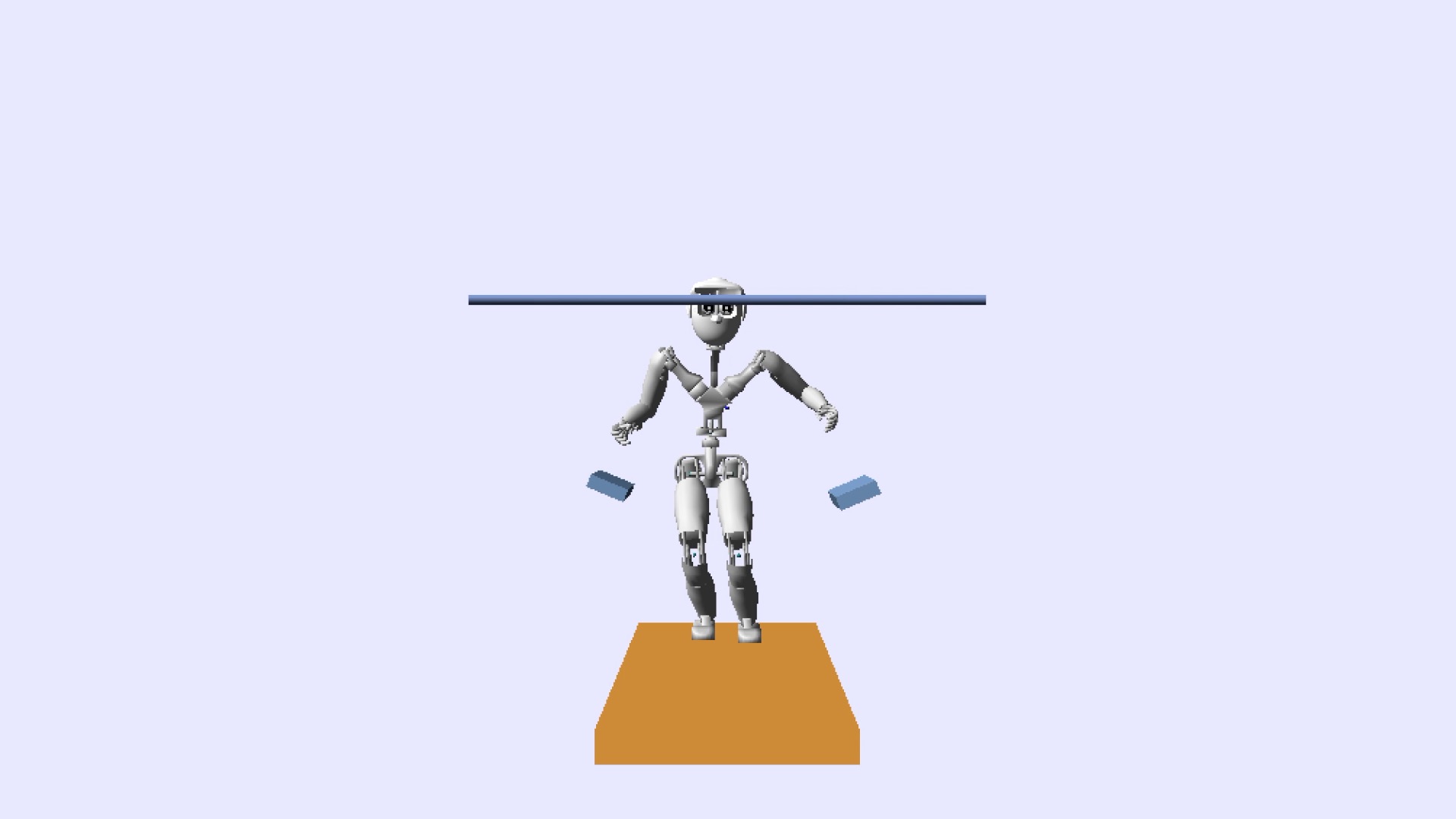}%
			\includegraphics[width=0.20\linewidth, trim={20cm 0cm 25cm 4cm}, clip]{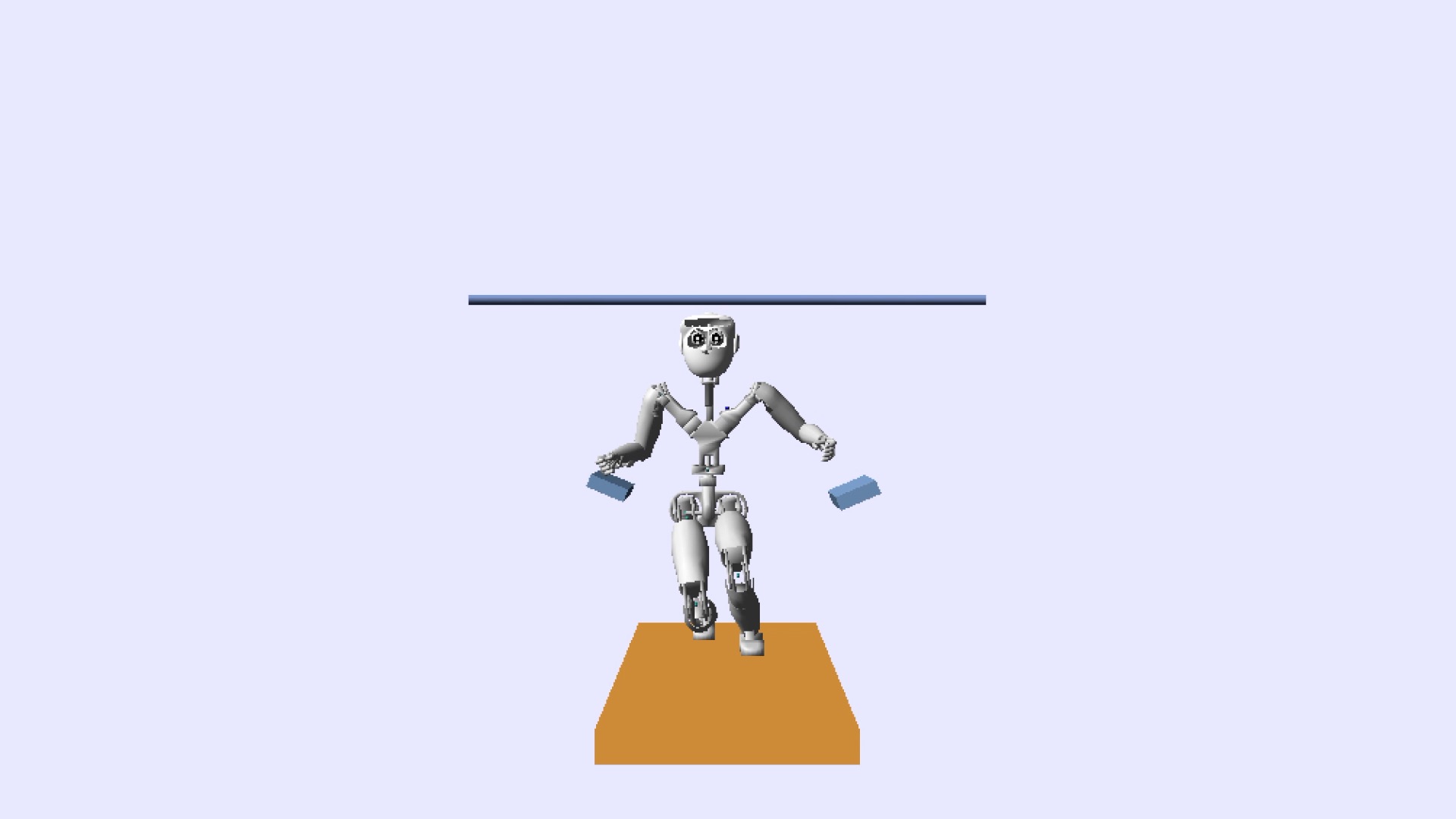}%
			\includegraphics[width=0.20\linewidth, trim={20cm 0cm 25cm 4cm}, clip]{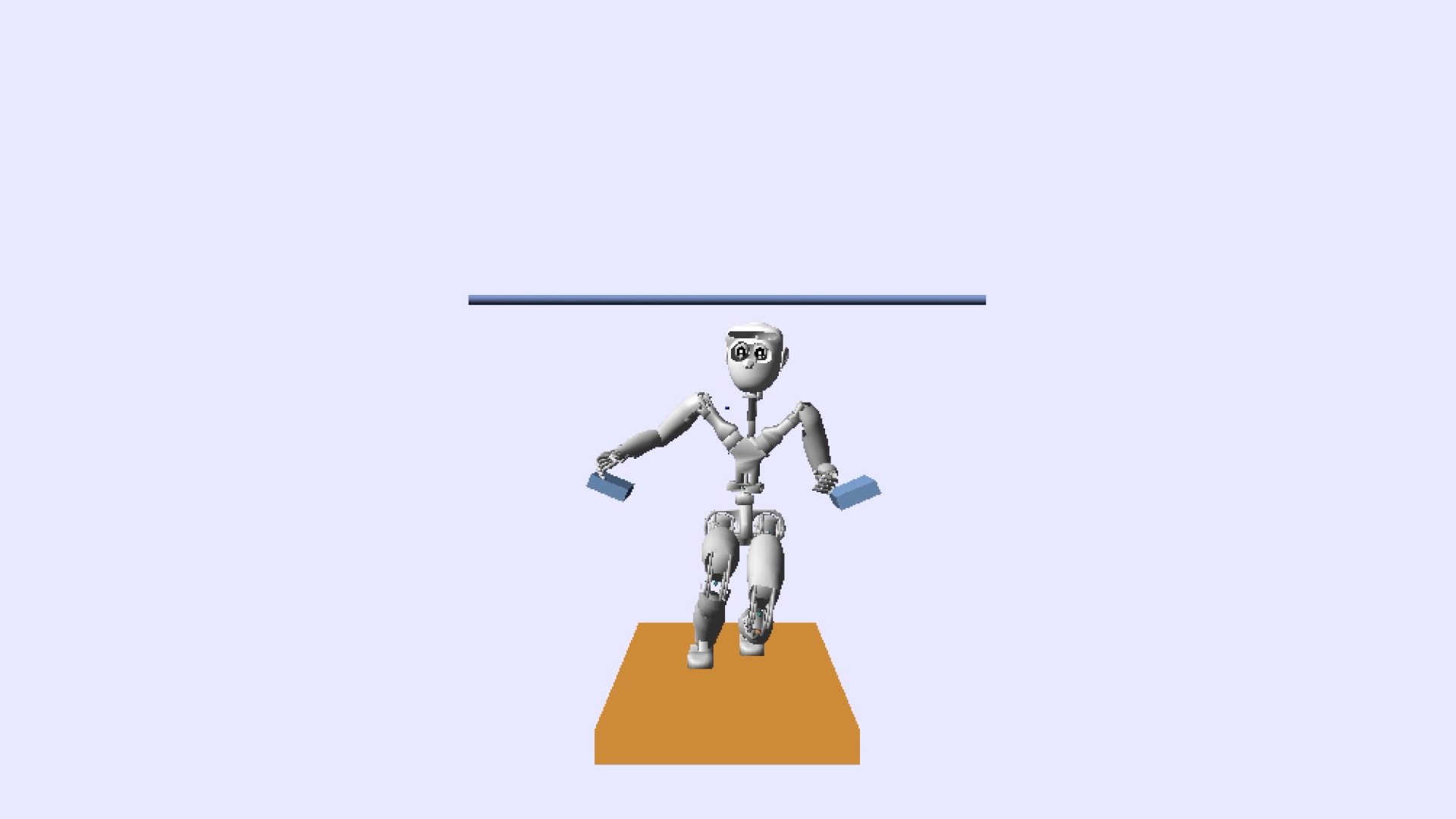}%
			\includegraphics[width=0.20\linewidth, trim={20cm 0cm 25cm 4cm}, clip]{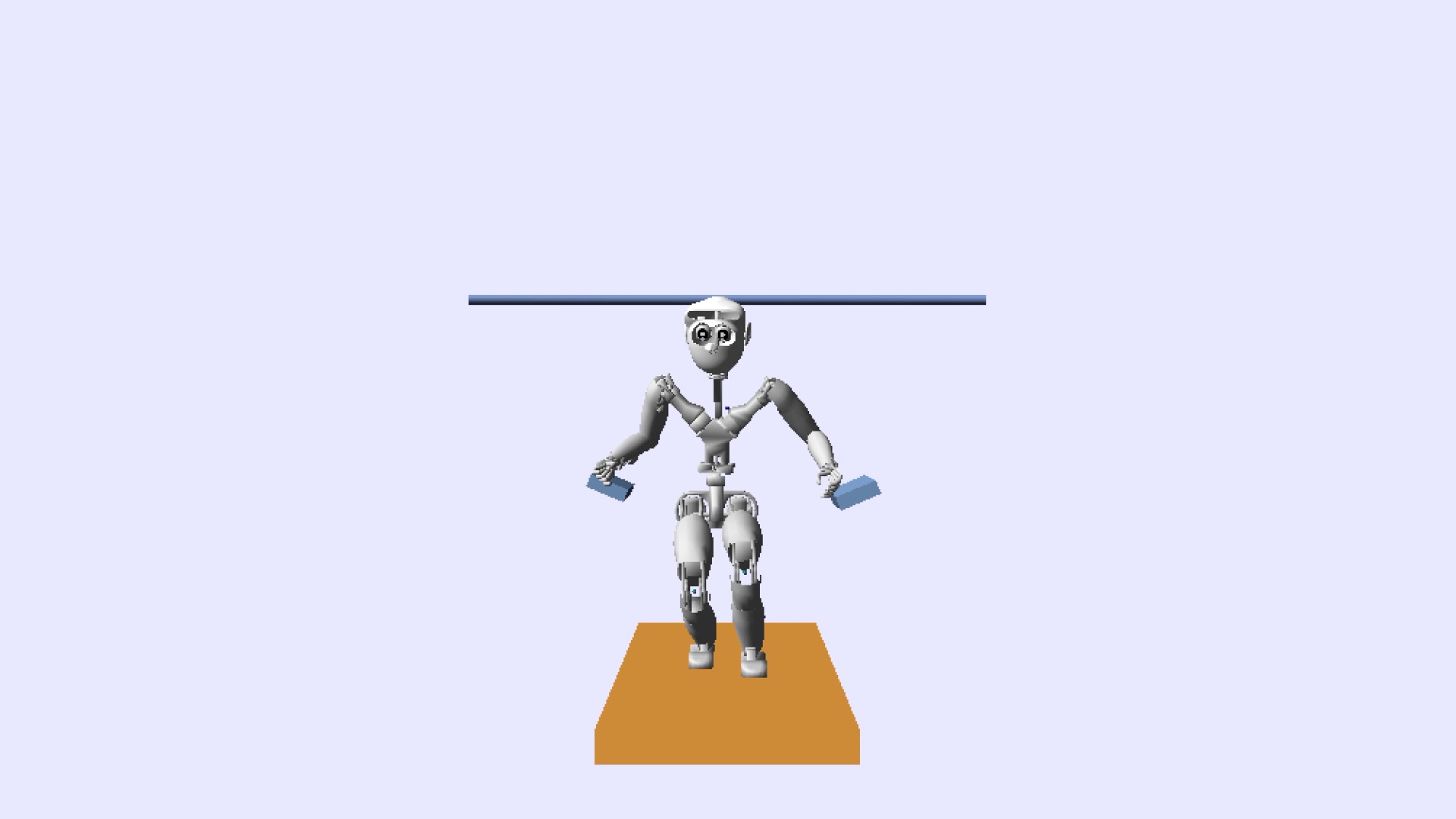}%
			\includegraphics[width=0.20\linewidth, trim={20cm 0cm 25cm 4cm}, clip]{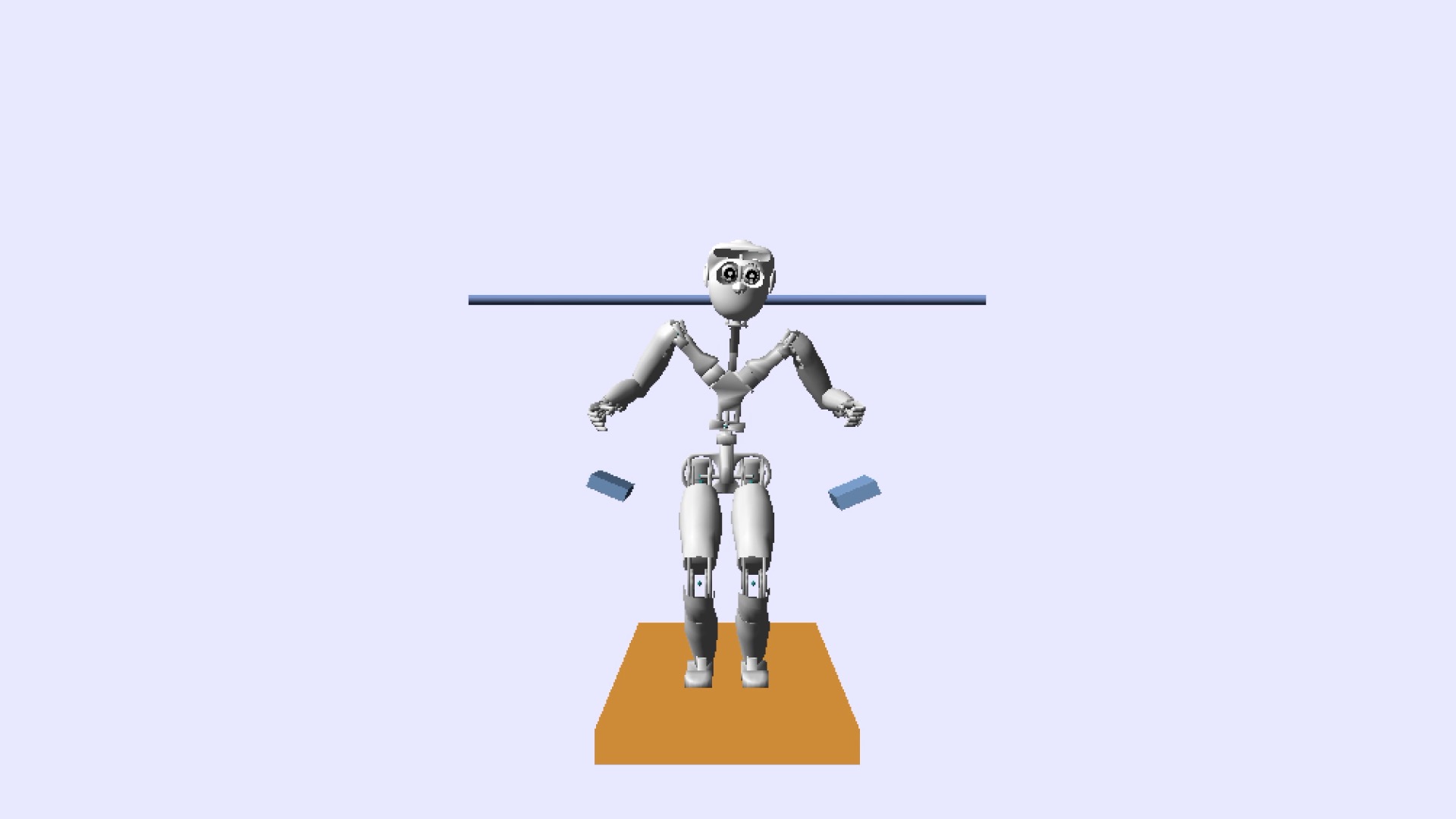}%
		}\\
		\vspace{0.4cm}
		\includegraphics[width=\textwidth]{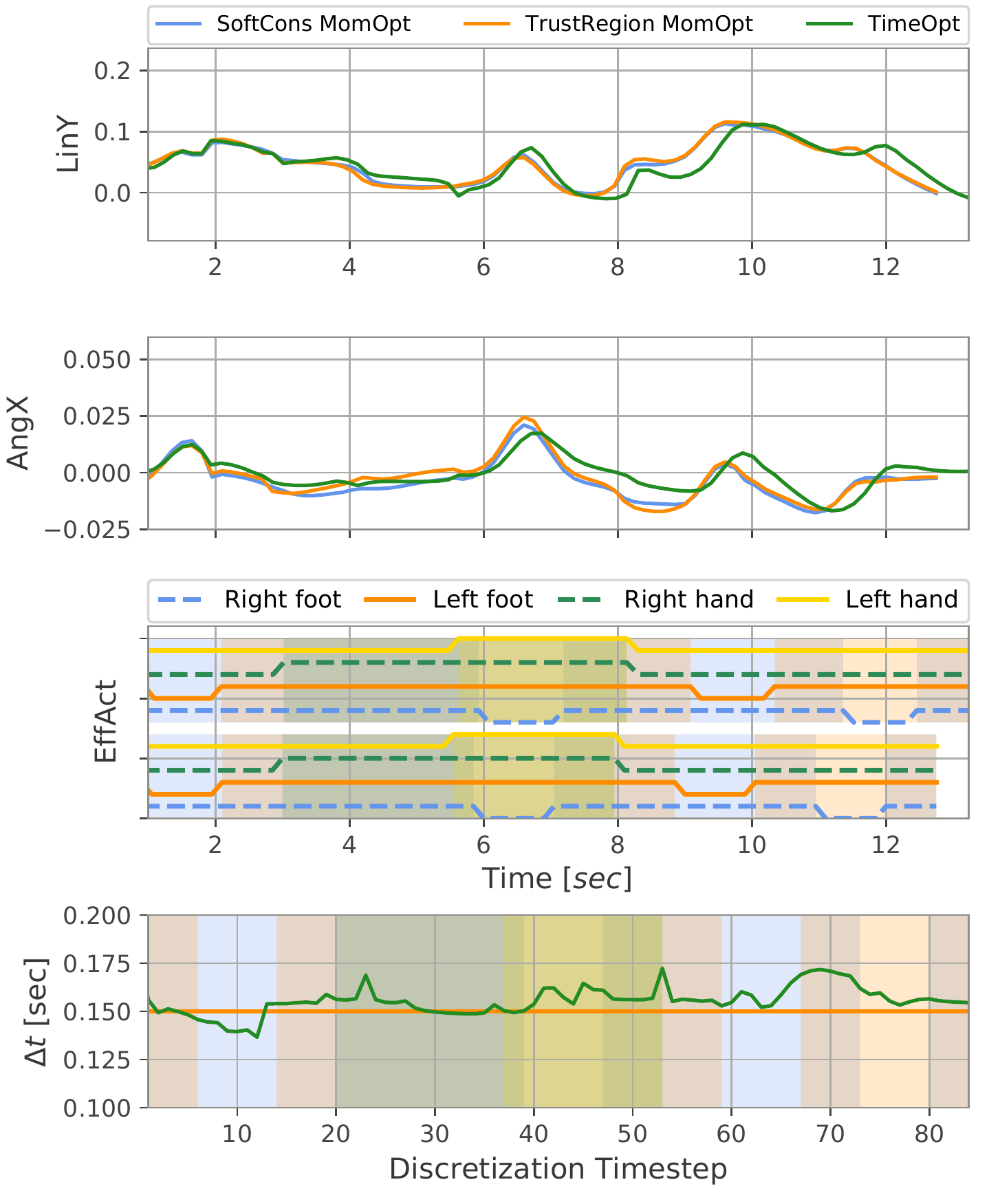}
		\caption[]%
		{{\small This figure shows optimization results for the walking under a bar using hands motion optimizing only momentum (blue) with soft constraint heuristic, with trust region heuristic (orange) and including time optimization (green). EffAct plot shows endeffector activations for time optimization (top) and momentum optimization with soft and trust region constraints (bottom).}}    
		\label{fig3:handing_motion}
	\end{subfigure}
	\caption[]
	{\small This figure shows examples of a walking motion and walking under a bar using hands motion, and its corresponding momentum trajectories normalized by robot mass. EffAct plots show the activation of each endeffector over time. Vertical colored bars helps us match endeffector activations over time with bottom plots displaying optimal duration of each time discretization. The number of time discretizations is fixed, e.g. in momentum optimization, a time horizon and a timestep duration are used to select equally time-spaced timesteps; in time optimization, the number of time discretizations is the same, but the timestep duration changes.} 
	\label{fig:experimental_results}
\end{figure*}

We have tested the algorithm in several multi-contact scenarios, including walking on an uneven terrain (Fig. \ref{fig2:walking_motion}), walking under a bar using also hands (Fig. \ref{fig3:handing_motion}), and a walking motion with low friction coefficient (Fig. \ref{fig4:lowfriction_motion}). The resulting motions, reproduced in time agreement with the optimized plans, are visible in the attached video https://youtu.be/ZGhSCILANDw.
%
%%%%%%%%%%%%%%%%%%%%%%%%%%%%%%%%%%%%%%%%%%%%%%%%%%%%
%%%%%%%%%%%%%%%%%%%%%%%%%%%%%%%%%%%%%%%%%%%%%%%%%%%%
%
\subsection{Walking on uneven terrain} The first motion (Fig. \ref{fig2:walking_motion}) has been built so that double support time after a single support on left foot is short, while double support time after single support on right foot is longer. It then allows to see the effect of double support duration on momentum optimization.
Stairs are also close to each other, such that legs require momentum to be lifted up, and minimum jerk trajectories that guide this kinematic motion have not been well tuned such that the effect of it is more visible. We test three scenarios: fixed time momentum optimization (blue), time optimization with no constraints on the total duration of the motion (orange) and time as an optimization variable with fixed total duration (green)

As visible in Fig. \ref{fig2:walking_motion} (Momentum Optimization, in blue), the angular momentum in the Y direction has peaks at short time double supports, while during longer double supports less momentum is distributed along a longer timespan. For the optimization of both time and momentum (orange), timestep discretizations are included in the optimization, without a fixed total duration. As can be seen in the bottom plot, timestep durations are increased during the initial short time double supports. The optimization thus automatically distributes the angular momentum in the Y direction without peaks. The last optimization (fixed time horizon and momentum optimization, in green), is also capable of adapting the timestep durations, however, to be comparable to the original motion, the time horizon is fixed to the same value as in the first experiment. Here, we can also see the tendency of increasing the timestep durations at the short time double supports, while reducing time spent at the beginning and end of the motion, and when walking on the straight line. This also allows to decrease angular momentum in the Y direction, which was the goal of the optimization. 
We note (not shown), that the momentum coming from the kinematic optimizer can be tracked by the dynamic optimization.

These experiments show how time can be used by the optimizer to reduce the overall momentum of the system and result in potentially easier to execute plans. For example, in an extreme case, where the available motion time makes the peaks of only the momentum optimized plan (in blue) much higher, the endeffector wrenches required to realize such a motion would also be higher. A different perspective on this could be a scenario with limitations on the endeffector wrench (such as walking under low friction) that would require a slower execution of the motion to make it feasible.
\subsection{Walking under a bar using hands} 
In this example, we compare both relaxation methods (trust region and soft constraint) using fixed time
(Fig. \ref{fig3:handing_motion}). We see that both relaxations lead to very similar solutions. This suggests both methods
are similarly applicable, however in our current implementation the soft constraint relaxation
was always significantly faster than the trust region one.
For this specific problem, we also notice that time optimization brings marginal changes to the motion, which highlights that having more endeffectors available to control the motion can compensate the lack of timing adaptation.
\subsection{Walking with low friction coefficient} 
\begin{figure}[h]
	\centering
	\drawvideo{5}{80}{%
		\includegraphics[width=0.20\linewidth, trim={20cm 4cm 25cm 0cm}, clip]{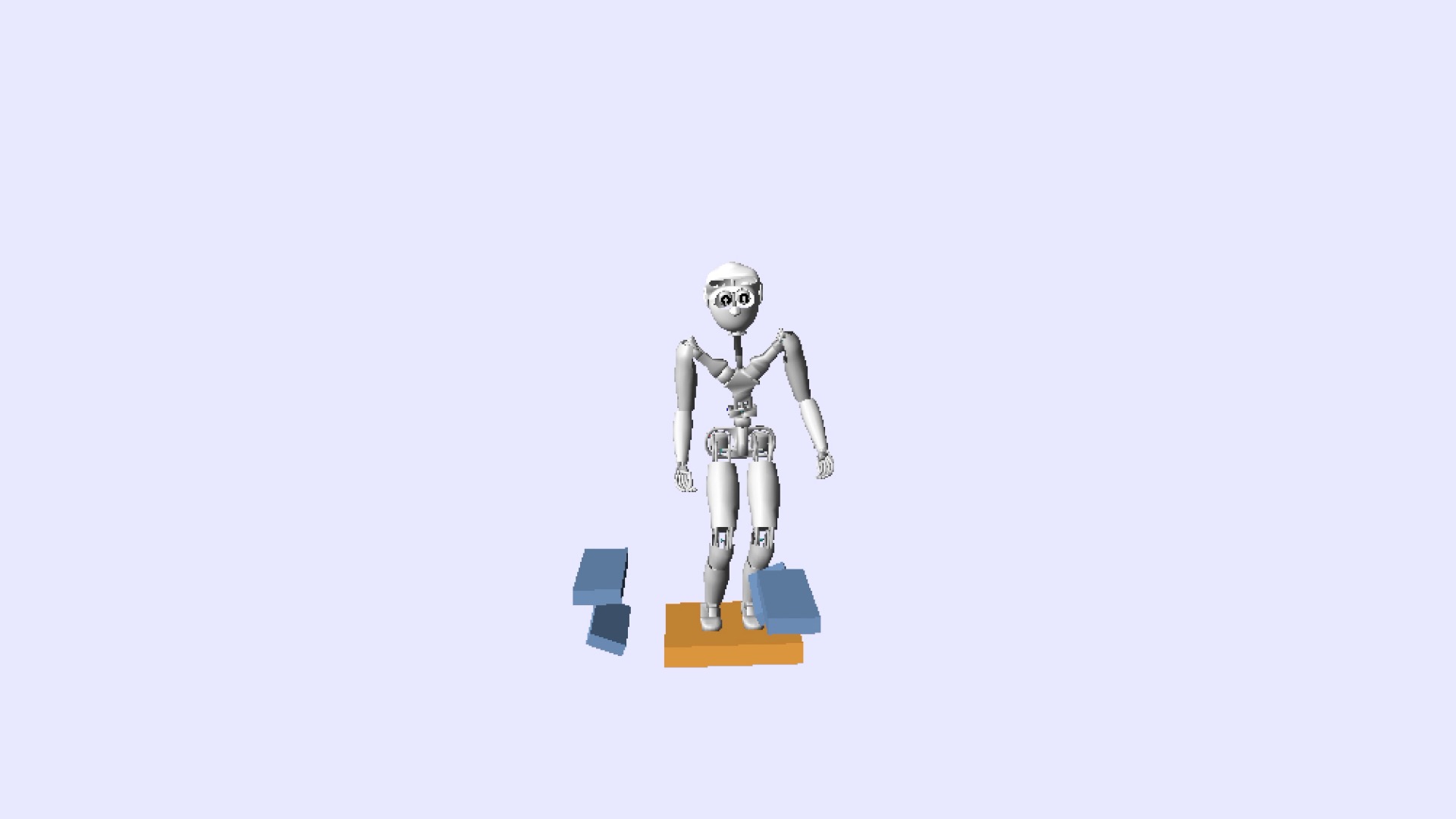}%
		\includegraphics[width=0.20\linewidth, trim={20cm 4cm 25cm 0cm}, clip]{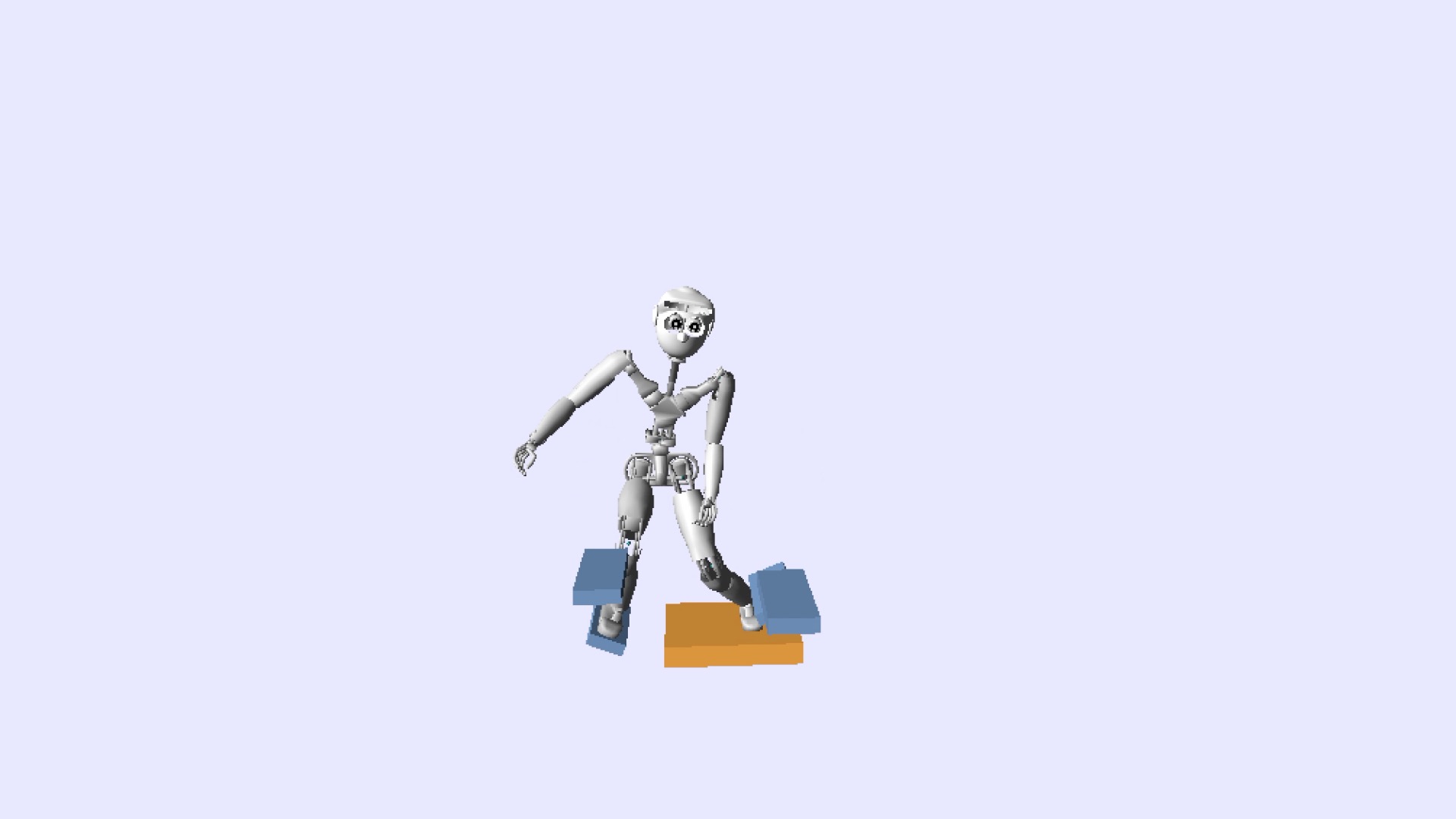}%
		\includegraphics[width=0.20\linewidth, trim={20cm 4cm 25cm 0cm}, clip]{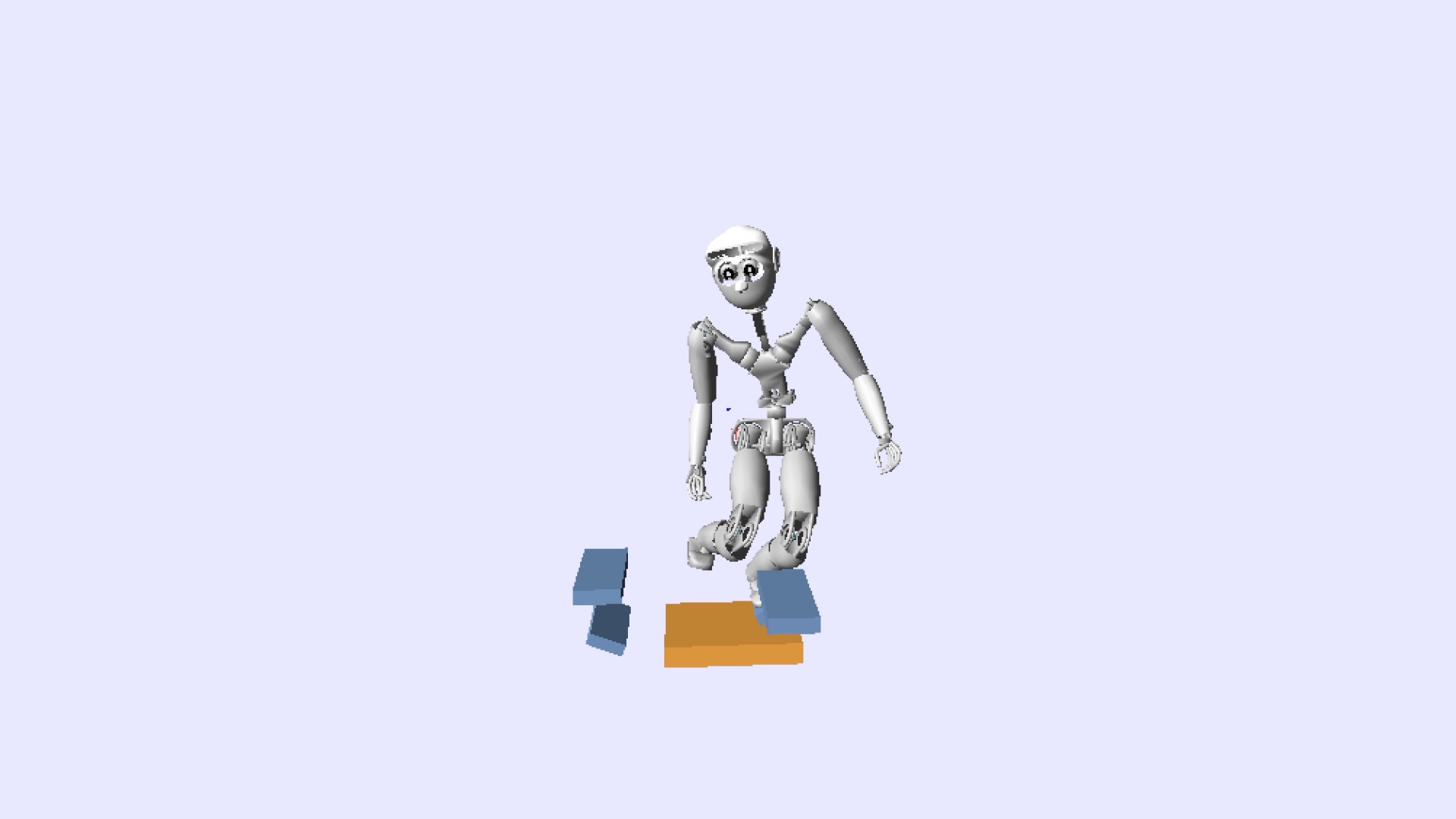}%
		\includegraphics[width=0.20\linewidth, trim={20cm 4cm 25cm 0cm}, clip]{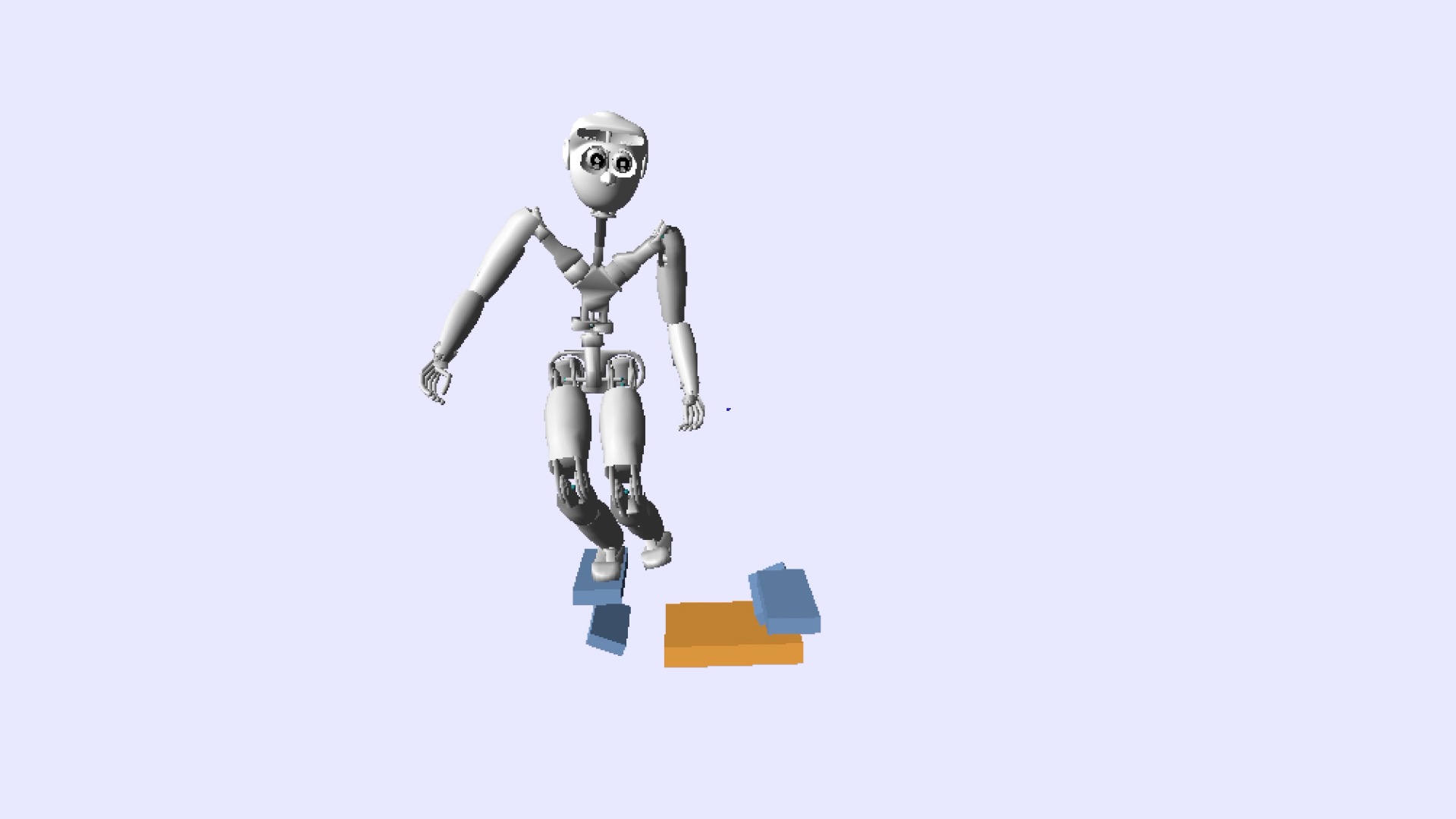}%
		\includegraphics[width=0.20\linewidth, trim={20cm 4cm 25cm 0cm}, clip]{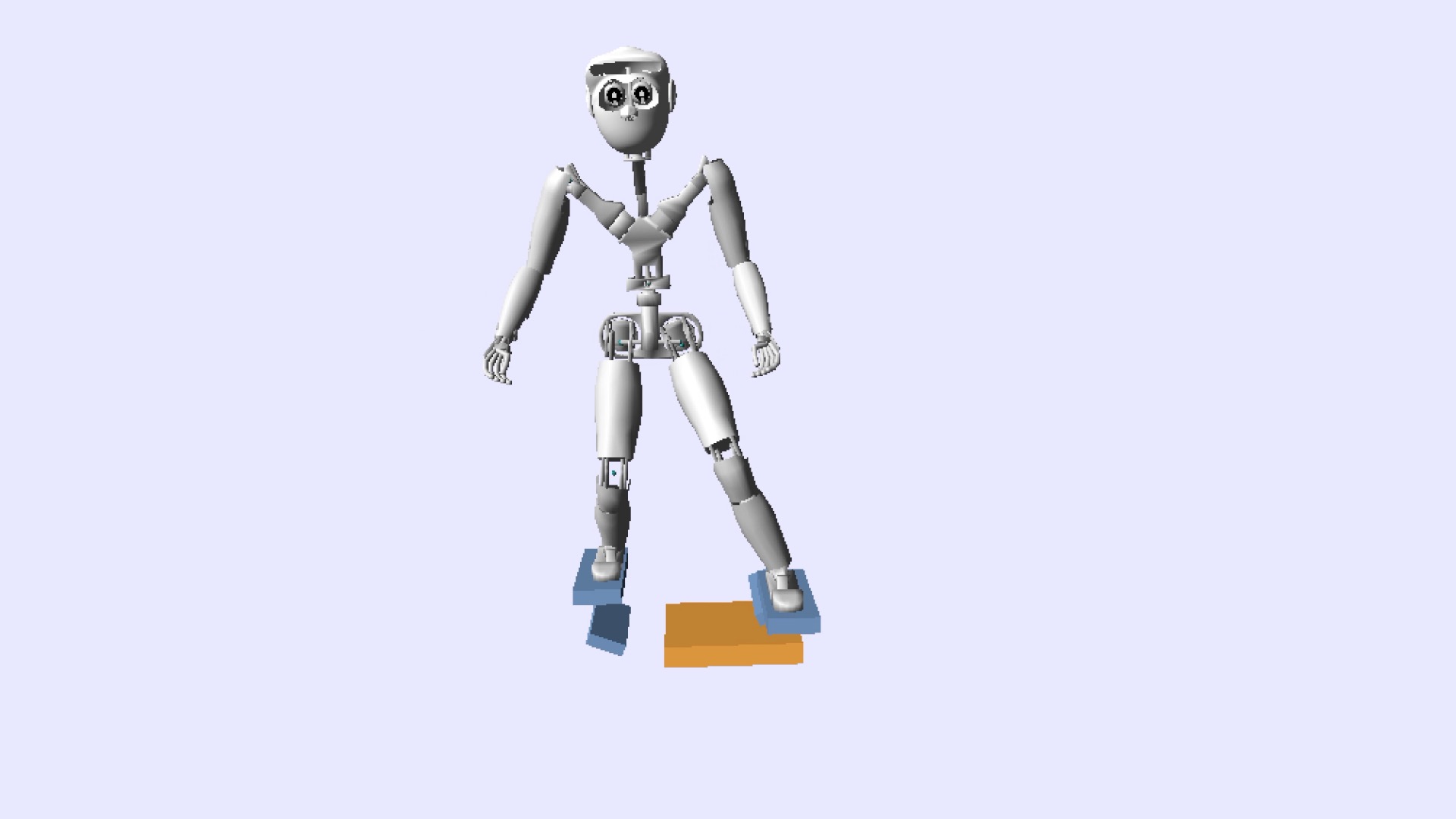}%
	}\\
    \vspace{0.4cm}
    \includegraphics[width=0.48\textwidth]{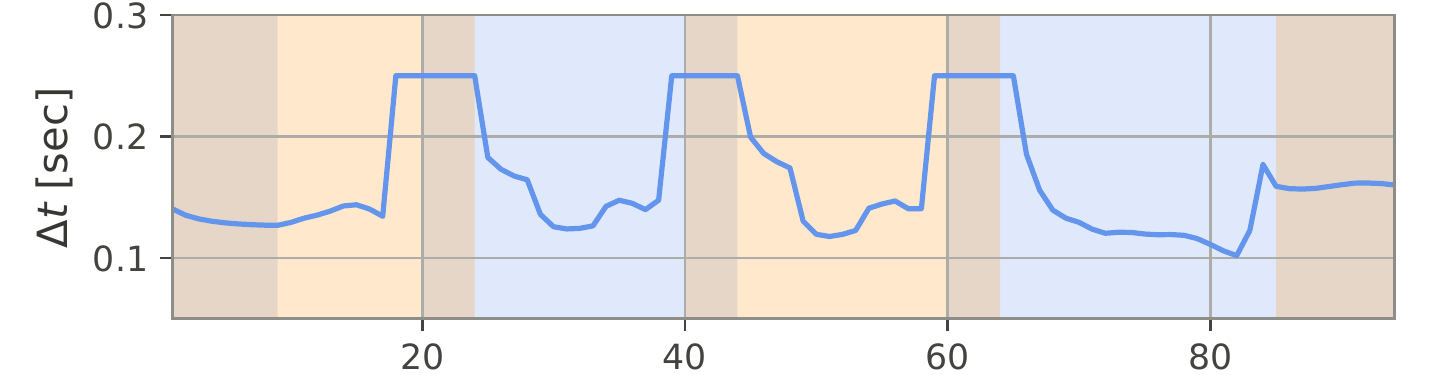}
	\caption{\small This figure shows optimization results for the walking under low friction motion. Bottom plot shows optimal duration of each timestep discretization. Vertical colored bars represent endeffector activations. Right foot activation is shown in blue, and left foot activation in orange. Double supports show both colors.}
	\label{fig4:lowfriction_motion}
\end{figure}

In this example (Fig. \ref{fig4:lowfriction_motion}), time optimization was critical to find a dynamically feasible motion. In this example, the friction coefficient is 0.4, the original time horizon is around 10 sec with a timestep duration of 0.1 sec. Under these conditions, the motion cannot be realized without optimizing time. Time optimization allows to automatically increase the time horizon (to around 16 sec) and the time available during double supports (timestep durations hitting its limit at 0.25 sec), which generates a physically realizable motion. Another alternative to generate the motion could be, as in the previous case, using hand contacts if available.
\subsection{Discussion on time optimization} As shown in the previous experiments, including time as an optimization variable is useful, and depending on the kind of motion it can really produce much lower cost solutions or even make them feasible which is an infinite improvement. However, it increases the dimensionality of the problem to be solved and consequently the time required to solve it.

\subsubsection{Limitations of the convex approximations} The problem at hand is nonconvex and thus hard to solve. The proposed heuristics help us reduce the effort required to find a solution by searching in a convex space an approximate solution (because of the constraint violations); however, the approach has also limitations. For instance, the trust region method could start with a bad initial guess, and the trust region built around it could render the local solution non optimal or even worse, it might overly restrict the problem making it primal infeasible. In the case of soft-constraint approximations, the penalty affects competing objectives, namely the amount of constraint violation and conditioning of the problem, which leads to problems such as having low constraint violation but poor problem conditioning or high constraint violation with faster convergence, therefore a reasonable tradeoff needs to be found. Source code provides more detail on the implementation and heuristics used in our experiments to speed up convergence, to find a solution realiably and robustly, and on how we build and refine trust region constraints and soft-constraint penalties.
\subsubsection{Time complexity} Table \ref{table:time_complexity} shows information about the optimal problems being solved and time required to find a solution. Among the parameters being shown are the time horizon, the number of timesteps, number of variables, linear equality and inequality constraints, second order cones, size and nonzeros of the KKT matrix, and finally the time required to solve the problem. $M1_{\textrm{time}}$ is the motion under low friction coefficient with time adaptation, while $M1$ is the same motion but with normal friction coefficient because time is fixed. $M2$ is the motion for walking under a bar using hands without time adaptation and $M2_{\textrm{time}}$ with time adaptation.
\begin{table}
	\begin{center}
		\begin{tabular}{|l|c|c|c|c|}
			\hline
			& $M1$ & $M1_{\textrm{time}}$ & $M2$ & $M2_{\textrm{time}}$ \\
			\hline
			Horizon [sec]     & 9.5      & 9.5      & 12.8     & 12.8    \\
			Timesteps          & 95       & 95       & 85        & 85      \\
			Variables            & 6808   & 7568   & 8969    & 9649   \\
			Lin. equalities     & 855     & 1425   & 765       & 1275   \\
			Lin. inequalities  & 2210    & 2400  & 3298     & 3468   \\
			SO cones           & 4393   & 4488   & 5876     & 5961   \\
			Size KKT             & 31838 & 33833 & 42412   & 44197  \\
			Nnz KKT             & 83322 & 87698 & 112090 & 116006 \\
			\hline
			time [sec]          & 0.829  & 5.632  & 1.082   & 5.469    \\
			\hline
		\end{tabular}
		\caption{\small Time complexity for optimization with/without time.}
		\label{table:time_complexity}
	\end{center}
	%\vspace{-0.5cm}
\end{table}
We can observe that the timings required to solve a dynamics optimization problem are in the order of a second, which is double the time required in \cite{ConvexModelMomentumDynamics}, because in this method, we also refine the solution after the relaxation, which allows us to directly penalize and track momentum, which increases the time required to find a solution, however improves its quality. Our solving time is comparable to the one reported in \cite{TROCarpentier}, which corresponds to 1.23 sec for a motion duration of 8 sec using the centroidal wrench and 3.89 sec for the same motion using contact forces as control input, as in our case.
Note that the time taken by the optimization is always faster than the duration of the motion and that,
with appropriate warm start of the optimizer, receding horizon control is attainable.

The solve time for a motion that includes time optimization is larger because more iterative refinements are required to converge to a good threshold of constraint violation. However, it could be further sped up doing only a few iterations of time optimization and then fixing time discretizations; increasing the accepted approximation error tolerance used as convergence criteria; or warm-starting a time optimization with a fixed-time dynamic optimization. The computation time is still lower than the plan time horizon, what makes it possible to run the algorithm online (the next plan can be computed, while the current one is being executed).
\subsubsection{Constraint violations} Table \ref{table:constraint_violations} compares the average error in the center of mass, linear and angular momentum, computed integrating forces and torques with the original model and the approximate values in our relaxed formulation. As shown, errors on center of mass and linear momentum are marginal, while error on the angular momentum is small. The values shown are in an absolute scale. If they are compared to values shown in Figs. \ref{fig:experimental_results}-\ref{fig4:lowfriction_motion}, they are also small, because values in the figures are normalized by the robot mass.
\begin{table}
	\begin{center}
		\begin{tabular}{|l|c|c|}
			\hline
			& $M1_{\textrm{time}}$ & $M2_{\textrm{time}}$ \\
			\hline
			Center of Mass        &  1.187e-09  & 1.206e-07      \\
			Linear Momentum    &  3.954e-09  & 1.022e-06      \\
			Angular Momentum &  0.007  & 0.001      \\
			\hline
		\end{tabular}
		\caption{\small Amount of constraint violation.}
		\label{table:constraint_violations}
	\end{center}
	%\vspace{-0.5cm}
\end{table}
%
%%%%%%%%%%%%%%%%%%%%%%%%%%%%%%%%%%%%%%%%%%%%%%%%%%%%
%
\section{CONCLUSION} \label{sec:conclusion}
We have presented two convex relaxation methods for the optimization of the centroidal dynamics and
motion timing of a legged robot. Our approach is efficiently solvable and could therefore be used in receding horizon control. Moreover, the convex relaxation deviates only marginally from the original dynamics.
Our approach has not been yet tested on a real robot, but this is the step coming.


%%%%%%%%%%%%%%%%%%%%%%%%%%%%%%%%%%%%%%%%%%%%%%%%%%%%

\bibliographystyle{ieeetr}
\bibliography{references}

\end{document}